\documentclass[a4paper,11pt]{article}
\usepackage[utf8]{inputenc}

\usepackage[T1]{fontenc}
\usepackage{charter}
\usepackage[expert]{mathdesign}

\usepackage{titling}
\usepackage{cite}
\usepackage{amsmath}
\usepackage{amssymb}
\usepackage{amsfonts}
\usepackage{graphicx}
\usepackage{sectsty}
\usepackage[DIV=12,headinclude=true,footinclude=false]{typearea}
\usepackage{tikz}
\usepackage{booktabs}
\usepackage{mathrsfs}
\usepackage{todonotes}
\usepackage[printonlyused]{acronym}
\usepackage[colorlinks=true]{hyperref}
\usepackage{dsfont}
\usepackage[ruled,vlined,linesnumbered]{algorithm2e}
\usepackage{cleveref}
\usepackage{subcaption}
\usepackage{rotating}
\usepackage{mathtools}
\usepackage{listings}
\usepackage{setspace}

\makeatletter
\AtBeginDocument{%
	\renewcommand*{\AC@hyperlink}[2]{%
		\begingroup
		\hypersetup{hidelinks}%
		\hyperlink{#1}{#2}%
		\endgroup
	}%
}
\makeatother

\newacro{ANN}{artificial neural network}
\newacro{ARD}{automatic relevance determination}
\newacro{CUDA}{Compute Unified Device Architecture}
\newacro{GCD}{greatest common divisor}
\newacro{GP}{Gaussian process}
\newacro{GPN}{Gaussian process neuron}
\newacro{GPU}{graphics processing unit}
\newacro{CDF}{cumulative density function}
\newacro{PDF}{probability density function}
\newacro{MCMC}{Markov chain Monte Carlo}
\newacro{iid.}{independent and identically distributed}
\newacro{HMC}{Hamiltonian Monte Carlo}
\newacro{SE}{squared exponential}
\newacro{CNN}{convolutional neural network}
\newacro{RNN}{recurrent neural network}
\newacro{ELBO}{evidence lower bound}
\newacro{DFT}{discrete Fourier transform}

\renewcommand{\vec}[1]{\boldsymbol{#1}}

\newcommand{\N}{\mathcal{N}}

\newcommand{\setR}{\mathbb{R}}

\renewcommand{\P}{\mathrm{P}}
\newcommand{\Q}{\mathrm{Q}}

\newcommand{\GP}{\mathcal{GP}}
\newcommand{\ie}{i.e.\ }

\newcommand{\wrt}{w.r.t.\ }
\newcommand{\SE}{\mathrm{SE}}
\renewcommand{\d}{\mathrm{d}}

\renewcommand{\|}{\, | \,}

\newcommand{\diag}{\mathrm{diag}}

\newcommand{\idmatrix}{\mathds{1}}

\newcommand{\abs}[1]{\left| #1 \right|}

\renewcommand{\L}{\mathcal{L}}
\newcommand{\E}{\mathrm{E}}

\newcommand{\teq}{\triangleq}

\newcommand{\tr}{\mathrm{tr}}
\newcommand{\Cov}{\mathrm{Cov}}
\newcommand{\Var}{\mathrm{Var}}

\renewcommand{\O}{\mathcal{O}}

\newcommand{\Gp}{\ac{GP}}
\newcommand{\Gps}{\acp{GP}}
\newcommand{\GPN}{\ac{GPN}}
\newcommand{\GPNs}{\acp{GPN}}
\newcommand{\PDF}{\ac{PDF}}
\newcommand{\PDFs}{\acp{PDF}}

\newcommand{\iid}{\ac{iid.}}
\newcommand{\HMC}{\ac{HMC}}
\providecommand{\Se}{}
\renewcommand{\Se}{\ac{SE}}
\newcommand{\Ard}{\ac{ARD}}
\newcommand{\ANN}{\ac{ANN}}
\newcommand{\ANNs}{\acp{ANN}}

\newcommand{\CNNs}{\acp{CNN}}

\newcommand{\RNNs}{\acp{RNN}}
\newcommand{\CDF}{\ac{CDF}}

\newcommand{\chol}{\mathrm{chol}}

\newcommand{\ELBO}{\ac{ELBO}}

\newcommand{\mutilde}{\widetilde{\mu}}
\newcommand{\muhat}{\widehat{\mu}}

\newcommand{\Sigmatilde}{\widetilde{\Sigma}}
\newcommand{\Sigmahat}{\widehat{\Sigma}}
\newcommand{\Ptilde}{\widetilde{\P}}

\newcommand{\Xhat}{\widehat{X}}
\newcommand{\Xtilde}{\widetilde{X}}

\newcommand{\Khat}{\widehat{K}}
\newcommand{\Ktilde}{\widetilde{K}}

\let\textcite\cite

\begin{document}

\thanksmarkseries{arabic}
\author{Sebastian Urban\thanks{Technical University Munich, surban@tum.de}, Marcus Basalla\thanks{Technical University Munich, marcus.basalla@tum.de}, Patrick van der Smagt\thanks{AI Research, Volkswagen Group}}
\title{Gaussian Process Neurons \\ Learn \\ Stochastic Activation Functions}
\date{November 2017}

\begin{titlingpage}
\maketitle
\begin{abstract}
We propose stochastic, non-parametric activation functions that are fully learnable and individual to each neuron.
Complexity and the risk of overfitting are controlled by placing a Gaussian process prior over these functions.
The result is the Gaussian process neuron, a probabilistic unit that can be used as the basic building block for probabilistic graphical models that resemble the structure of neural networks.
The proposed model can intrinsically handle uncertainties in its inputs and self-estimate the confidence of its predictions.
Using variational Bayesian inference and the central limit theorem, a fully deterministic loss function is derived, allowing it to be trained as efficiently as a conventional neural network using mini-batch gradient descent.
The posterior distribution of activation functions is inferred from the training data alongside the weights of the network.

The proposed model favorably compares to deep Gaussian processes, both in model complexity and efficiency of inference.
It can be directly applied to recurrent or convolutional network structures, allowing its use in audio and image processing tasks.

As an preliminary empirical evaluation we present experiments on regression and classification tasks, in which our model achieves performance comparable to or better than a Dropout regularized neural network with a fixed activation function.
Experiments are ongoing and results will be added as they become available.
\end{abstract}
\end{titlingpage}	

\tableofcontents
\newpage

\section{Introduction}

We introduce a neural activation function that can be learned completely from training data alongside the weights of the neural network.
The number of constraints on the form of the learnable functions should be kept as low as possible to allow the highest amount of flexibility.
However, we want the neural network to be trainable using stochastic gradient descent, thus we require the activation function to be at least continuously differentiable.
Naturally by increasing the flexibility of a model the risk of overfitting is also increased.
Hence, to keep that risk in check, we apply a prior over the space of activation functions.
We choose a \Gp{} with a zero mean and \Se{} covariance function for that prior, since it encourages smooth functions of small magnitude.
This probabilistic treatment transforms a neuron into a probabilistic unit, which we call \GPN{}.
Consequently a neural network built from \GPNs{} becomes a probabilistic graphical model, which can intrinsically handle training and test data afflicted with uncertainties and estimate the confidence of its own predictions.

Since a \Gp{} is a non-parametric model it introduces dependencies between samples into the model.
Consequently, even after learning the network weights, the predictions on a test input directly depend on the training samples and inference has to be performed using Monte Carlo methods.
Since this is impractical, we introduce the parametric \GPN{}, an auxiliary model that applies methods from sparse \Gp{} regression to represent its activation function using a set of trainable parameters.
The resulting model approximation can be trained using stochastic gradient descent by sampling the gradient.

Then we will combine ideas from fast Dropout and sparse variational \Gps{} to make a \emph{non-parametric} \GPN{} network trainable using variational Bayesian inference.
The result is a fully deterministic loss function that eliminates the need to sample the gradient and thus makes training more efficient.
Although the model is treated fully probabilistically, the loss function retains the functional structure of a neural network, making it directly applicable to network architectures such as \acp{RNN} and \acp{CNN}.

The proposed model shares some commonalities with deep Gaussian processes but is more economical in terms of model parameters and considerably more efficient to train as will be discussed in \cref{sec:gpn_vs_deep_gp}.
An overview of the family of introduced models and their relationships is shown in \cref{fig:gpn_overview} at the end of this paper.

\subsection{Related Work}

Since the dawn of neural network research the most commonly used activation functions were the logistic function and other sigmoid functions.
Sigmoid refers to ``S''-shaped functions, for example the hyperbolic tangent (tanh).
This choice of activation functions was not seriously challenged by researchers (except for special purpose applications), until recently when \textcite{Nair2010} introduced the rectified linear unit (ReLU), a neuron with an activation function that is linear for positive inputs and zero for negative inputs.
\textcite{NIPS2012_4824} showed that ReLUs produce significantly better results on image recognition tasks using deep networks than the common sigmoid-shaped activation functions.

This achievement led to a wave of follow-up research in activation functions specifically tailored to deep networks.
While the ReLU solved the problem of vanishing gradients for positive values, it completely cut off the gradient for negative ones; thus once a neuron enters the negative regime (either through initialization or during training) for most samples, no training signal can pass through it.
To mitigate this problem \textcite{maas2013rectifier} introduced the leaky ReLU, which is also linear for negative values but with a very small, although non-zero, slope; for positive values it behaves like the ReLU.
Soon after \textcite{he2015delving} demonstrated that it is advantageous to make the slope of the negative part of the leaky ReLU an additional parameter of each neuron.
This parameter was trained alongside the weights and biases of the neural network using gradient descent.
A \ac{CNN} using these so-called parametric ReLUs was the first to surpass human-level performance on the ImageNet classification task \cite{deng2009imagenet}.


It is thus natural to ask if even more flexible activation functions are beneficial.
This question was answered affirmative by \textcite{agostinelli2014learning} on the CIFAR-10 and CIFAR-100 benchmarks \cite{krizhevsky2014cifar}.
The authors introduced piecewise linear activation functions that have an arbitrary (but fixed) number of points where the function changes its slope.
These points and the associated slopes are inferred from training data by stochastic gradient descent.

Instead of having a fixed parameter for the negative slope of the ReLU, \textcite{xu2015empirical} introduced stochasticity into the activation function by sampling the value for the slope with each training iteration from a fixed uniform distribution.
\textcite{clevert2015fast} and \textcite{klambauer2017self} replaced the negative part of ReLUs with a scaled exponential function and showed that, under certain conditions, this leads to automatic renormalization of the inputs to the following layer and thereby simplifies the training of the neural networks.
Furthermore \textcite{urban2015neural} introduced a parametric activation function to smoothly interpolate the operation a neuron performs on its inputs between addition and multiplication, thus allowing the most appropriate operation to be determined during training.

Nearly fully adaptable activation functions have been proposed by \textcite{Eisenbach:2017}.
The authors use a Fourier basis expansion to represent the activation function; thus with enough coefficients any (periodic) activation function can be represented.
The coefficients of this expansion are trained as network parameters using stochastic gradient descent.
Similarly, \textcite{scardapane2017kafnets} also use a basis expansion, but with a set of Gaussian kernels that are equally distributed over a preset input range.

\clearpage
\section{Prerequisites}

\subsection{Tensor Slicing}\label{sec:slicing}
In this work the need arises to slice tensors along one or more dimension.
A star ($\star$) will be used in place of the index to select all indices of that dimension.
Let us provide a few examples.
Given a matrix $X \in \setR^{N \times M}$ the notation $X_{i\star} \in \setR^M$ denotes the $i$-th row of $X$.
Similarly $X_{\star j} \in \setR^N$ denotes the $j$-th column of $X$.

This can also be extended to tensors and the star can be used multiple times.
For example, consider the tensor $A \in \setR^{N_1 \times N_2 \times N_3 \times N_4}$.
Here $A_{i\star j \star} \in \setR^{N_2 \times N_4}$ denotes the matrix that is obtained by fixing the first dimension of $A$ to $i$ and the third dimension to $j$.

\subsection{Gaussian Processes}
A \acl{GP}~\cite{williams1996gaussian} describes a distribution over scalar-valued functions $f(\vec{x})$ with multivariate inputs $\vec{x}$.
Consider a matrix $X \in \setR^{N \times D} $ of $N$ input values with $D$ dimensions.
Any finite number of function values $f(X_{i\star})$, $i \in \{1, 2, \dots N\}$, has a joint multivariate normal distribution, which is defined by the mean function $m: \setR^D \to \setR$ and covariance function $k: \setR^D \times \setR^D \to \setR$ of the \Gp{}.
Let $\vec{f} \in \setR^N$ be the vector of function values given by $f_i \teq f(X_{i\star})$.
If the function $f$ follows a \Gp{},
\begin{equation}\label{eq:gp_f}
f(\vec{x}) \sim \GP(m(\vec{x}), k(\vec{x}, \vec{x'})) \,, 
\end{equation}
this means that the function values $\vec{f}$ are normally distributed,
\begin{equation}\label{eq:gp_f_points}
\P(\vec{f}) \sim \N(\vec{f} \| \vec{m}, K(X,X)) \,, 
\end{equation}
with mean and covariance determined by the inputs $X$,
\begin{equation}
m_i         \teq m(X_{i\star}) \,, 
\quad\quad
K(X,X)_{ij} \teq k(X_{i\star}, X_{j\star}) \,.
\end{equation}
This implies that $k(\vec{x}, \vec{x}')$ must by symmetric and positive-definite to be a valid covariance function.

\subsubsection{Gaussian Process Regression}\label{sec:gp_regression}
\Gps{} can be used to perform regression by conditioning the \Gp{} on a set of observed function values.
Consider a function $f(\vec{x})$ distributed according to~\eqref{eq:gp_f}.
Given a finite set of observation points $X_{i\star}$, $i \in \{1, 2, \dots, N\}$, with corresponding (noisy) observed values $\vec{y} = (y_1, y_2, \dots, y_N)^T$ where $y_i = f(X_{i\star}) + \epsilon$ with $\epsilon \sim \N(0, \sigma_n^2)$ the conditional distribution of the predicted function values $f(\vec{x^*})$ at some test points $\vec{x^*}$ is another \Gp{},
\begin{equation}\label{eq:gp_f_reg}
f(\vec{x^*}) \| X, \vec{y} \,\sim\,  \GP(m^*(\vec{x^*}), k^*(\vec{x^*}, \vec{x^*}')) \,, 
\end{equation}
with predictive mean and covariance functions given by
\begin{subequations}\label{eq:gp_f_reg_mean_var}
\begin{align}
m^*(\vec{x^*})              &\teq m(\vec{x^*}) + K(\vec{x^*}, X) \, [K(X,X) + \sigma_n^2 \idmatrix]^{-1} \, (\vec{y} - \vec{m}) \,, \\
k^*(\vec{x^*}, \vec{x^*}')  &\teq k(\vec{x^*}, \vec{x^*}') - K(\vec{x^*}, X) \, [K(X,X) + \sigma_n^2 \idmatrix]^{-1} \, K(X, \vec{x^*}') \,,
\end{align}
\end{subequations}
where $K(\vec{x^*}, X)_i \teq k(\vec{x^*}, X_{i\star})$,  $K(X, \vec{x^*}')_i \teq k(X_{i\star}, \vec{x^*}')$, $m_i \teq m(X_{i\star})$ and $\idmatrix$ is the identity matrix.
This follows directly from the equations for the mean and covariance of a conditional, multivariate normal distribution, since any finite number of points from the \Gp{} must be consistent with it.

\clearpage
\section{Gaussian Process Neurons}\label{sec:gpn}

\begin{figure}[tb]
\centering
\begin{tikzpicture}
    \tikzstyle{unit}=[circle,red,very thick,draw,minimum size=14pt,text=black,font=\footnotesize]
	\tikzstyle{annot} = [text width=4em, text centered]

	\node[unit,label=270:{$X^{l-1}_{12}$}] (X12) at (2.2,0) {};
	\node[unit,label=270:{$X^{l-1}_{11}$}] (X11) at (1.0,0) {};
	\node[unit,label=270:{$X^{l-1}_{13}$}] (X13) at (3.4,0) {};
	\node[unit,label=270:{$X^{l-1}_{21}$}] (X21) at (5.0,0) {};
	\node[unit,label=270:{$X^{l-1}_{22}$}] (X22) at (6.2,0) {};
	\node[unit,label=270:{$X^{l-1}_{23}$}] (X23) at (7.4,0) {};
	\node[unit,label=270:{$X^{l-1}_{31}$}] (X31) at (9.0,0) {};
	\node[unit,label=270:{$X^{l-1}_{32}$}] (X32) at (10.2,0) {};
	\node[unit,label=270:{$X^{l-1}_{33}$}] (X33) at (11.4,0) {};

	\node[unit,label=100:{$A^l_{11}$}] (V1) at (2.2,1.6) {};
	\node[unit,label=100:{$A^l_{21}$}] (V2) at (6.2,1.6) {};
	\node[unit,label=100:{$A^l_{31}$}] (V3) at (10.2,1.6) {};

	\node[unit,label=100:{$F^l_{11}$}] (F1) at (2.2,3.2) {};
	\node[unit,label=100:{$F^l_{21}$}] (F2) at (6.2,3.2) {};
	\node[unit,label=100:{$F^l_{31}$}] (F3) at (10.2,3.2) {};

	\node[unit,label={$X^l_{11}$}] (Y1) at (2.2,4.8) {};
	\node[unit,label={$X^l_{21}$}] (Y2) at (6.2,4.8) {};
	\node[unit,label={$X^l_{31}$}] (Y3) at (10.2,4.8) {};

	\path[thick,->] (X11) edge (V1);        
	\path[thick,->] (X12) edge (V1);        
	\path[thick,->] (X13) edge (V1);        
	\path[thick,->] (X21) edge (V2);        
	\path[thick,->] (X22) edge (V2);        
	\path[thick,->] (X23) edge (V2);        
	\path[thick,->] (X31) edge (V3);        
	\path[thick,->] (X32) edge (V3);        
	\path[thick,->] (X33) edge (V3);        

	\path[thick,->] (V1) edge (F1);        
	\path[thick,->] (V2) edge (F2);        
	\path[thick,->] (V3) edge (F3);        

	\path[ultra thick] (F1) edge (F2);        
	\path[ultra thick] (F2) edge (F3);        

	\path[thick,->] (F1) edge (Y1);        
	\path[thick,->] (F2) edge (Y2);        
	\path[thick,->] (F3) edge (Y3);        

	\draw[gray] (4.2, -1.4) -- (4.2, 5.5);
	\draw[gray] (8.2, -1.4) -- (8.2, 5.5);
	\node[annot] at (2.2,-1.2) {sample 1};
	\node[annot] at (6.2,-1.2) {sample 2};
	\node[annot] at (10.2,-1.2) {sample 3};
	\node[annot] at (1.0,0.8) {$W^l$};
	\node[annot] at (5.0,0.8) {$W^l$};
	\node[annot] at (9.0,0.8) {$W^l$};
	\node[annot] at (1.6,4.8) {$\vec{\sigma}^l$};
	\node[annot] at (5.6,4.8) {$\vec{\sigma}^l$};
	\node[annot] at (9.6,4.8) {$\vec{\sigma}^l$};
	\node[annot] at (-0.2,0.0) {\emph{inputs}};	
	\node[annot] at (-0.2,1.6) {\emph{activation}};	
	\node[annot] at (-0.2,3.2) {\emph{response}};
	\node[annot] at (-0.2,4.8) {\emph{output}};
\end{tikzpicture}
\caption{
A \GPN{} within layer $l$ with three inputs is shown for three samples as a probabilistic graphical model.
The inputs are the outputs of the previous layer and represented by the random variables $X^{l-1}_{sm}, s \in \{1, 2, 3\}, m \in \{1, 2, 3\}$.
The activations for each sample are represented by the random variables $A^l_{s}, s \in \{1, 2, 3\}$, and depend deterministically on the inputs.
The responses $F^l_{\star}$ are a \Gp{} over the samples conditioned on the activations; in the figure the \Gp{} is represented by a Markov random field shown as an undirected connection between all samples.
The outputs are represented by the random variables $X^l_{s}$, $s \in \{1, 2, 3\}$.
}
\label{fig:gpn}
\end{figure}
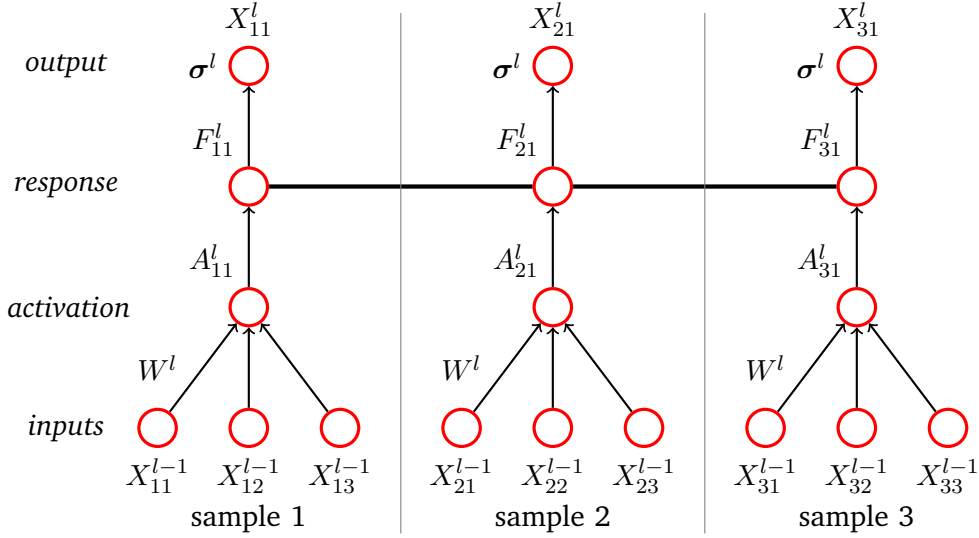

We introduce \aclp{GPN}~(GPNs) in the context of feed-forward neural networks, \ie we assume that \GPNs{} are arranged in layers and the outputs of one layer are fed as inputs to the next layer.
A \GPN{} is a probabilistic unit that receives multiple inputs and computes an output distribution conditioned on the values of its inputs.
Given multiple input samples the output samples of a \GPN{} become correlated, \ie the output distribution is \emph{not} \iid{} over the samples.
A probabilistic graphical model corresponding to a single \GPN{} within a layer with three inputs and three samples is shown in \cref{fig:gpn}.

Let $l \in \{1, \dots, L\}$ be the layer index and $N_l$ the number of \GPNs{} in that layer.
Further let $s \in \{1, \dots S\}$ denote a data point (sample).
The activations $A^l_{sn}$ depend deterministically on the inputs $X^{l-1}_{s\star}$ via the weights, 
\begin{equation} 
A^l_{s\star} =  X^{l-1}_{s\star} \, W^l 
\end{equation}
The response, \ie result of applying the activation function in a conventional neuron, follows a \Gp{} prior,
\begin{equation}\label{eq:gpn_value}
f (a) \sim \GP (0, k^\lambda_{\SE{}} (a, a')) \,,
\end{equation}
with zero mean function.
This \Gp{} has \emph{scalar} inputs and uses the standard \Se{} covariance function,
\begin{equation}
k^\lambda_{\SE{}}(a,a') = \exp\!\left(-\frac{(a-a')^2}{2\,\lambda^2}\right) \,,
\end{equation}
with lengthscale $\lambda$.
For a finite set of samples this implies
\begin{equation}\label{eq:F_given_X}
\P(F^l_{\star n} \| X^{l-1}) = \N\bigl(\vec{0}, K^l_n(A^l_{\star n}, A^l_{\star n})\bigr) = \N\bigl(\vec{0}, K^l_n(X^{l-1} \, W^l_{\star n} , X^{l-1} \, W^l_{\star n} )\bigr) \,
\end{equation}
with the covariance matrix $[ K^l_n(\vec{a}, \vec{a'}) ]_{ij} \teq k^{\lambda^l_n}_{\SE{}}(a_i, a_j)$.
The outputs follow the responses with additive Gaussian noise,
\begin{equation}\label{eq:X_given_F}
\P(X^l_{\star n} \| F^l_{\star n}) = \N\bigl(F^l_{\star n}, (\sigma_n^l)^2 \idmatrix\bigr) \,.
\end{equation}
Marginalizing over the distributions of $A^l$ and $F^l$ leads to the conditional distribution of the output given the inputs
\begin{equation}\label{eq:gpn_output_marginal}
\P(X^l_{\star n} \| X^{l-1}) = \N(X^l_{\star n} \| \vec{0}, \Ktilde^{\lambda^l_n}_{W^l_{\star n}}(X^{l-1}, X^{l-1}) + (\sigma^l_n)^2 \idmatrix )    \,,
\end{equation}
with $[\Ktilde^\lambda_{\vec{w}}(X, X')]_{ij} \teq k^\lambda_{\vec{w}}(X_{i\star}, X'_{j\star})$ and the \GPN{} covariance function
\begin{equation}\label{eq:gpn_cov_fn}
k^\lambda_{\vec{w}}(\vec{x}, \vec{x'}) = \exp\!\left(-\frac{\left[\sum_m w_m (x_m - x'_m)\right]^2}{2\,\lambda^2}\right) \,.
\end{equation}
Note that~\eqref{eq:gpn_output_marginal} has the same structure as a deep \Gp{}, however the proposed covariance function~\eqref{eq:gpn_cov_fn} differs from the \Ard{} covariance function, which is used in deep \Gps{}, by multiplying the weights with the inputs \emph{before} taking the square.
We will show in \cref{sec:gpn_vs_deep_gp} that this difference results in a model that is considerably more efficient to train.
Furthermore, note that the lengthscale $\lambda$ is redundant to the magnitude $\abs{\vec{w}}^2$ of the weight vector and we can set $\lambda=1$ for now.
However, since it will be useful for a parametric version of the model introduced later, we keep it.

\subsection{Non-Parametric Training and Inference}\label{sec:nonp_training}
Performing training and inference in a non-parametric \GPN{} model is computationally intensive, as we will show now.
Let the random variable $\Xhat^l_{sn}$ denotes the $n$-th \GPN{} in layer $l$ corresponding to training sample $s$.
Test samples are represented by the random variables $\Xtilde^l_{sn}$.
The layer $l=0$ corresponds to the inputs and its values are observed for both the training and test samples.
The top-most layer $l=L$ represents the outputs.
For defining the distributions occurring in the model, it is convenient to concatenate the training and test samples of all inputs and \GPNs{}, \ie $X^l_{\star n} = \left[ (\Xhat^l_{\star n})^T, (\Xtilde^l_{\star n})^T \right]^T$.
Analogous to standard neural networks, \GPN{} layers can be stacked to form a multi-layer feed-forward network.
The joint probability of training and test samples in such a \GPN{} stack given the inputs $X^0$ is 
\begin{equation}\label{eq:gpn_stack}
\P(\{X\}_1^L \| X^0) = \prod_{l=1}^L \prod_{n=1}^{N_l} \P(X_{\star n}^{l} \| X^{l-1} ) \,.
\end{equation}
with $\P(X_{\star n}^{l} \| X^{l-1})$ given by~\eqref{eq:gpn_output_marginal} and using the notation shorthand $\{\bullet\}_1^L \teq \bullet^1, \bullet^2, \dots, \bullet^L$.

\GPN{} models are a hybrid between parametric and non-parametric models.
They contain parameters $\theta \teq \{\theta^1, \theta^2, \dots, \theta^L\}$ with $\theta^l \teq \{W^l, \vec{\sigma}^l, \vec{\lambda}^l \}$ consisting of the weights, variances and lengthscales for each layer.
However, their predictions on test samples also depend directly on the training samples since the \Gp{} forms a Markov random field over \emph{all} samples.
This does not occur in conventional \ANNs{}, in which the predictions for all samples are \acl{iid.}.

Thus making predictions on the test set using \GPNs{} consists of two steps.
One, obtain an estimate for the parameters $\theta$ by maximizing their likelihood $\P_{\theta}(\Xhat^L \| \Xhat^0)$ on the training set.
Two, compute the predictive distribution $\P_{\theta}(\Xtilde^L \| \Xtilde^0, \Xhat^0, \Xhat^L)$ or sample from it.
Both steps require all latent variables $X^1, \dots, X^{L-1}$ to be marginalized out.
Unfortunately due to the occurrence of $X^{l-1}$ in the covariance matrix in~\eqref{eq:gpn_output_marginal} analytic integration is intractable.
This is because the inverse of the covariance matrix appears in the \PDF{} of a normal distribution and thus the dependency on $X^{l-1}$ is highly non-linear and analytic calculation of the integral is not feasible.

It is possible to estimate the likelihood and its derivatives by sampling from the model using \HMC{} sampling.
Using this estimate of the gradient \wrt the parameters the likelihood can be maximized iteratively using stochastic gradient ascent.
In the same fashion \HMC{} sampling can be used to sample from $\P_{\theta}(\Xtilde^L \| \Xtilde^0, \Xhat^0, \Xhat^L)$ and thus obtain predictions on the test set.
However, this approach is not very attractive when \GPNs{} should be used in place of conventional neurons, since \HMC{} sampling is considerably more expensive than the standard backpropagation algorithm and the non-parametric nature requires us to consult the whole training set for each prediction, thus limiting the scalability of the model to datasets of arbitrary size.

\clearpage
\section{Parametric Gaussian Process Neurons}\label{sec:parametric_gpn}
The issues described in \cref{sec:nonp_training} arise from the inter-sample dependencies in the non-parametric \GPN{} model.
Consequently it is desirable to break them up, however it has to be ensured that the activation function of each \GPN{} is still consistent across all samples.
To make \Gps{} scalable to large datasets \textcite{quinonero2005unifying} proposed to find a sparse approximation of the training data and base the predicitons of the \Gp{} solely on it.
Here, we use the same method to make each \GPN{} parametric and thus its output independent of the training data.
Later we will show how this approach can be extended to make analytic marginalization of the latent variables tractable.

Three parameter vectors $\vec{V} \in \setR^{R}$, $\vec{U} \in \setR^{R}$ and $\smash{\vec{S} \in (\setR^{+})^{R}}$ are introduced per \GPN{}.
Their purpose is to explicitly parameterize $R$ points of the activation function representing the mode of the \Gp{} distribution.
For each virtual observation point $r \in \{1,2, \dots, R\}$ this parameterization consists of an inducing point $V_r$, corresponding to the activation of the observation, the target $U_r$, corresponding to the response given that activation, and the variance $S_r$ around the response.
Thus we assume that we are making observations of the activation function $f(a)$ given by~\eqref{eq:gpn_value}.
These observations are of the form
\begin{equation}
f(V_r) = U_r + \epsilon \quad \text{with } \epsilon \sim \N(0, S_r) \,, \quad r \in \{1, 2,\dots, R\} \,.
\end{equation}
In a \GPN{} stack each \GPN{} in has its own virtual observations, thus we add indices for the layer and \GPN{} within the layer to the virtual observation variables.
Consequently we obtain the tensors $V^l_{rn}$, $U^l_{rn}$ and $S^l_{rn}$ for the $r$-th inducing point, target and variance respectively of \GPN{} $n$ in layer $l$.

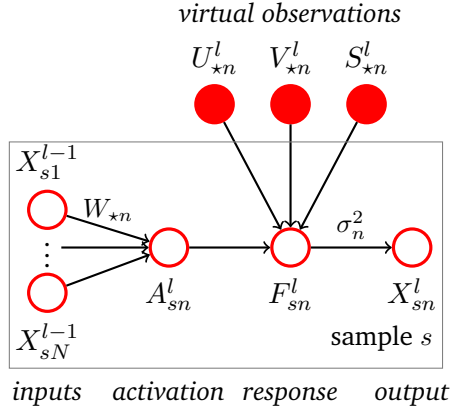
\begin{figure}
\centering
\begin{tikzpicture}
    \tikzstyle{unit}=[circle,red,very thick,draw,minimum size=14pt,text=black,font=\footnotesize]
	\tikzstyle{annot} = [font=\small]

	\node[unit,label=270:{$X^{l-1}_{sN}$}] (X11) at (0,1.4) {};
	\node[annot] (X12) at (0.0,2.0) {$\vdots$};
	\node[unit,label=90:{$X^{l-1}_{s1}$}] (X13) at (0,2.5) {};
	
	\node[unit,label=270:{$A^l_{sn}$}] (A1) at (1.6,2.0) {};

	\node[unit,label=270:{$F^l_{sn}$}] (F1) at (3.2,2.0) {};

	\node[unit,fill,label=90:{$U^l_{\star n}$}] (U) at (2.2,3.9) {};
	\node[unit,fill,label=90:{$V^l_{\star n}$}] (V) at (3.2,3.9) {};
	\node[unit,fill,label=90:{$S^l_{\star n}$}] (S) at (4.2,3.9) {};

	\node[unit,label=270:{$X^l_{sn}$}] (Y1) at (4.8,2.0) {};

	\path[thick,->] (X11) edge (A1);        
	\path[thick,->] (X12) edge (A1);        
	\path[thick,->] (X13) edge (A1);        

	\path[thick,->] (A1) edge (F1);        

	\path[thick,->] (V) edge (F1);        	
	\path[thick,->] (U) edge (F1);        	
	\path[thick,->] (S) edge (F1);        	

	\path[thick,->] (F1) edge (Y1);        

	\draw[gray] (-0.5,0.4) rectangle (5.2, 3.4);
	\node[annot] at (4.4,0.8) {sample $s$};
	\node[annot] at (0.8,2.5) {$W_{\star n}$};
	\node[annot] at (4.0,2.3) {$\sigma_n^2$};
	\node[annot,text height=1.5ex,text depth=.25ex,text centered] at (0.0,0.1) {\emph{inputs}};	
	\node[annot,text height=1.5ex,text depth=.25ex,text centered] at (1.6,0.1) {\emph{activation}};	
	\node[annot,text height=1.5ex,text depth=.25ex,text centered] at (3.2,0.1) {\emph{response}};
	\node[annot,text height=1.5ex,text depth=.25ex,text centered] at (4.8,0.1) {\emph{output}};
	\node[annot,text height=1.5ex,text depth=.25ex,text centered] at (3.2,5.1) {\emph{virtual observations}};
\end{tikzpicture}
\caption{
The auxiliary parametric representation of a GPN using virtual observation inducing points $V$, targets $U$ and variances $S$.
}
\label{fig:gpn_parametric}
\end{figure}

We use the virtual observations $V^l_{\star n}$ and $U^l_{\star n}$ as ``training'' points for a \Gp{} regression of the activation function evaluated at the ``test'' points $A^l_{\star n}$.
After performing the marginalizations over the activations and responses, as it was done to calculate~\eqref{eq:gpn_output_marginal}, we obtain for the distribution of the outputs given the inputs
\begin{equation}\label{eq:gpn_parametrized_y_given_x}
\P(X^l_{\star n} \| X^{l-1}) = \N(X^l_{\star n} \| \mu^{X^l}_{\star n}, \Sigma^{X^l}_{\star \star n} ) 
\end{equation}
with
\begin{subequations}\label{eq:gpn_parametrized_mu_sigma_y_sn}
\begin{align}
\mu^{X^l}_{\star n}          = &\, K^l_n(X^{l-1} W^l_{\star n}, V^l_{\star n}) \, [K^l_n(V^l_{\star n}, V^l_{\star n}) + \diag (S^l_{\star n})]^{-1} \, U^l_{\star n} \label{eq:gpn_parametrized_mu_y_sn} \\
\Sigma^{X^l}_{\star\star n}	 = &\, K^l_n(X^{l-1} W^l_{\star n}, X^{l-1} W^l_{\star n}) \, - \nonumber \\
					&\, K^l_n(X^{l-1} W^l_{\star n}, V^l_{\star n}) \, [K^l_n(V^l_{\star n}, V^l_{\star n}) + \diag (S^l_{\star n})]^{-1} K^l_n(V^l_{\star n}, X^{l-1} W^l_{\star n}) + (\sigma^l_n)^2 \, \idmatrix  \label{eq:gpn_parametrized_sigma_y_sn}
\end{align}
\end{subequations}
with the covariance matrices $K^l_n$ defined as before and $\diag (\vec{x})$ denoting a diagonal matrix with $\vec{x}$ on its diagonal.
The graphical model corresponding to that parametric version of the \GPN{} is shown in \cref{fig:gpn_parametric}.

As before we have for a layer
\begin{equation}\label{eq:gpn_parametrized_y_given_x_all_samples}
\P(X^l \| X^{l-1}) = \prod_{n=1}^{N_l} \P(X^l_{\star n} \| X^{l-1}) \,.
\end{equation}
Note that the inducing points are \emph{not} affected by the weights and always one dimensional, no matter what the dimensionality of the inputs $X^{l-1}$ is.
The parameters of a \emph{parametric \GPN{} layer} now include the virtual observations besides the weights, thus $\theta^l \teq \{W^l, \vec{\sigma}^l, \vec{\lambda}^l, V^l, U^l, S^l \}$ are the parameters of each layer that need to be estimated during training.
An overview of the notation used is provided in \cref{tab:gpn_notation}.

\begin{table}\centering
\begin{tabular}{ll} \toprule
\emph{symbol} 	& \emph{purpose} \\ \midrule
\multicolumn{2}{l}{\emph{model hyper-parameters:}} \\
$N_l$			& number of \GPNs{} in layer $l$ \vspace{2mm} \\
\multicolumn{2}{l}{\emph{random variables:}} \\
$Y^l_{sm}$		& input dimension $m$ of sample $s$ to \GPN{} layer $l$ \\
$A^l_{sn}$		& activation of \GPN{} $n$ in layer $l$ corresponding to sample $s$ \\
$F^l_{sn}$ 		& response of \GPN{} $n$ in layer $l$ corresponding to sample $s$ \\
$X^l_{sn}$		& output of \GPN{} $n$ in layer $l$ corresponding to sample $s$   \vspace{2mm} \\   
\multicolumn{2}{l}{\emph{model parameters:}} \\
$W^l_{nm}$		& weight from input $m$ to \GPN{} $n$ in layer $l$ \\
$(\sigma^l_n)^2$	& output variance of \GPN{} $n$ in layer $l$ \\
$\lambda^l_n$	& lengthscale of \GPN{} $n$ in layer $l$ \vspace{2mm} \\
\multicolumn{2}{l}{\emph{additional parameters for parametric \GPN{}:}} \\
$V^l_{rn}$		& virtual observation point $r$ of \GPN{} $n$ in layer $l$ \\
$U^l_{rn}$		& virtual observation target $r$ of \GPN{} $n$ in layer $l$ \\
$S^l_{rn}$		& virtual observation variance $r$ of \GPN{} $n$ in layer $l$ \\
\bottomrule 
\end{tabular}
\caption{Overview of the notation used for \GPNs{}. 
When \GPN{} layers are stacked a superscripted $l$ is used to denote the layer index and the output of one layer is the input to the next, thus $Y^l = X^{l-1}$.}
\label{tab:gpn_notation}
\end{table}
  
Since the model is fully parametric, after training all information about the training samples is stored in the parameters of the model and hence it is not necessary to keep training samples for prediction.
The number of parameters depends on the number of \GPNs{} and how many virtual observations are used per \GPN{}, but it is independent of the number of training samples.
This allows parametric \GPN{} networks to scale to datasets of arbitrary size, just like conventional neural networks do.

\subsection{Structure of Training Objective}
The standard method for training a model is to minimize a task-dependent loss $\mathscr{L}$ on the training set \wrt the model parameters $\theta$.
Here we consider losses of the form
\begin{equation}
\mathscr{L}(\theta) = \frac{1}{S} \sum_{s=1}^S L(X^L_{s\star}, T_{s\star}) \,,
\end{equation}
where $X^L_{s\star}$ is the prediction of the model for training sample $s$ and $T_{s\star}$ is the corresponding target.
The loss measure $L: \setR^D \times \setR^D \to \setR$ assigns a loss value to each sample based on the task-dependent difference between the model's prediction and the ground truth.
Since we are dealing with a probabilistic model that provides a predictive distribution $\P_{\theta}(X^L \| X^0)$, training is performed by considering the expectation of the loss, \ie the objective is to minimize
\begin{equation}
\mathscr{L}(\theta) = \E_{\P_{\theta}(X^L \| X^0)}\!\left[ \frac{1}{S} \sum_{s=1}^S L(X^L_{s\star}, T_{s\star}) \right] = \frac{1}{S} \sum_{s=1}^S \E_{\P_{\theta}(X^L_{s\star} \| X^0)} \!\left[ L(X^L_{s\star}, T_{s\star}) \right] \,.
\end{equation}
Note that here information is conveyed from sample to sample solely through the model parameters $\theta$.
Since the marginal $\P(X^l_{s n} \| X^{l-1}_{\star \star})$ of the multivariate normal~\eqref{eq:gpn_parametrized_y_given_x} with respect to sample $s$ depends only on $\mu^{X^l}_{s\star}$ and $\Sigma^{X^l}_{ss\star}$ and furthermore these quantities depend only on $X^{l-1}$ via $X^{l-1}_{s\star}$ as can be seen from~\eqref{eq:gpn_parametrized_mu_sigma_y_sn}, we observe that $\P(X^l_{s\star} \| X^{l-1}) = \P(X^l_{s\star} \| X^{l-1}_{s\star})$ for all layers $l$, and thus the objective becomes
\begin{equation}\label{eq:gpn_parametrized_loss}
\mathscr{L}(\theta) = \frac{1}{S} \sum_{s=1}^S \E_{\P_{\theta}(X^L_{s\star} \| X^0_{s\star})} \!\left[ L(X^L_{s\star}, T_{s\star}) \right] \,.
\end{equation}
It is possible to approximate this expectation by sampling from $\P_{\theta}(X^L_{s\star} \| X^0_{s\star})$, which is straightforward to do, since the conditionals $\P_{\theta}(X^l_{s\star} \| X^{l-1}_{s\star})$ are normal distributions and can be sampled sequentially for each layer $l$.
Thus a maximum likelihood estimate of the model parameters $\theta$ can by obtained by performing stochastic gradient descent using sampled gradients after applying the reparameterization trick~\cite{kingma2013auto} on~\eqref{eq:gpn_parametrized_loss}.

While stochastic training has been proven to find parameter optima that generalize better to unseen data samples, it also slows training down considerably because it leads to noisy estimates of the gradient, which, in turn, require the learning rate to be kept at least a magnitude lower compared to noise-free algorithms to prevent the trajectory in parameter space from becoming unstable.
To increase the speed of training and make it comparable with that of a conventional neural network, we will therefore demonstrate how to analytically calculate $\P_{\theta}(X^L_{s\star} \| X^0_{s\star})$ by propagating distributions through a parametric \GPN{} network in the following section.


\subsection{Central Limit Activations}

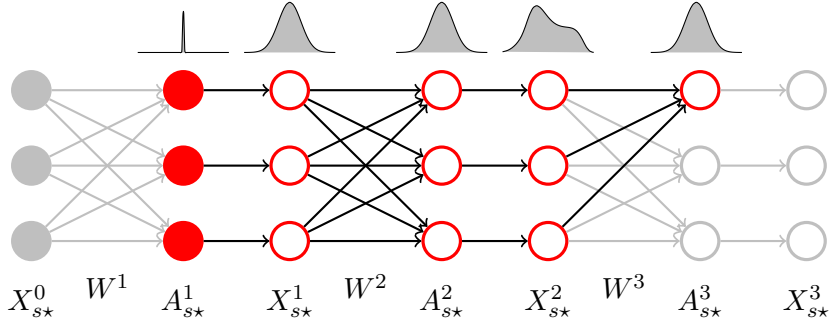
\begin{figure}[bt]
\centering
\begin{tikzpicture}
    \tikzstyle{unit}=[circle,red,very thick,draw,minimum size=14pt,text=black]
    \tikzstyle{par}=[circle,red,thick,draw,minimum size=8pt,text=black,fill]
	\tikzstyle{annot} = [text width=4em, text centered]	

	\def\la{0.0}
	\def\lb{2.0}
	\def\lc{3.4}
	\def\ld{5.4}
	\def\le{6.8}
	\def\lf{8.8}
	\def\lg{10.2}
	
	\node[unit,fill,lightgray] (X011) at (\la,0.5) {};
	\node[unit,fill,lightgray] (X012) at (\la,1.5) {};
	\node[unit,fill,lightgray] (X013) at (\la,2.5) {};

	\node[unit,fill] (A111) at (\lb,0.5) {};
	\node[unit,fill] (A112) at (\lb,1.5) {};
	\node[unit,fill] (A113) at (\lb,2.5) {};

	\node[unit] (X111) at (\lc,0.5) {};
	\node[unit] (X112) at (\lc,1.5) {};
	\node[unit] (X113) at (\lc,2.5) {};

	\node[unit] (A211) at (\ld,0.5) {};
	\node[unit] (A212) at (\ld,1.5) {};
	\node[unit] (A213) at (\ld,2.5) {};

	\node[unit] (X211) at (\le,0.5) {};
	\node[unit] (X212) at (\le,1.5) {};
	\node[unit] (X213) at (\le,2.5) {};

	\node[unit,lightgray] (A311) at (\lf,0.5) {};
	\node[unit,lightgray] (A312) at (\lf,1.5) {};
	\node[unit] (A313) at (\lf,2.5) {};

	\node[unit,lightgray] (X311) at (\lg,0.5) {};
	\node[unit,lightgray] (X312) at (\lg,1.5) {};
	\node[unit,lightgray] (X313) at (\lg,2.5) {};

	\path[thick,lightgray,->] (X011) edge (A111);        
	\path[thick,lightgray,->] (X011) edge (A112);        
	\path[thick,lightgray,->] (X011) edge (A113);        
	\path[thick,lightgray,->] (X012) edge (A111);        
	\path[thick,lightgray,->] (X012) edge (A112);        
	\path[thick,lightgray,->] (X012) edge (A113);        
	\path[thick,lightgray,->] (X013) edge (A111);        
	\path[thick,lightgray,->] (X013) edge (A112);        
	\path[thick,lightgray,->] (X013) edge (A113);        

	\path[thick,->] (A111) edge (X111);        
	\path[thick,->] (A112) edge (X112);        
	\path[thick,->] (A113) edge (X113);        

	\path[thick,->] (X111) edge (A211);        
	\path[thick,->] (X111) edge (A212);        
	\path[thick,->] (X111) edge (A213);        
	\path[thick,->] (X112) edge (A211);        
	\path[thick,->] (X112) edge (A212);        
	\path[thick,->] (X112) edge (A213);        
	\path[thick,->] (X113) edge (A211);        
	\path[thick,->] (X113) edge (A212);        
	\path[thick,->] (X113) edge (A213);        

	\path[thick,->] (A211) edge (X211);        
	\path[thick,->] (A212) edge (X212);        
	\path[thick,->] (A213) edge (X213);        

	\path[thick,->,lightgray] (X211) edge (A311);        
	\path[thick,->,lightgray] (X212) edge (A311);        
	\path[thick,->,lightgray] (X213) edge (A311);        

	\path[thick,->,lightgray] (X211) edge (A312);        
	\path[thick,->,lightgray] (X212) edge (A312);        
	\path[thick,->,lightgray] (X213) edge (A312);        

	\path[thick,->] (X211) edge (A313);        
	\path[thick,->] (X212) edge (A313);        
	\path[thick,->] (X213) edge (A313);        

	\path[thick,->,lightgray] (A311) edge (X311);        
	\path[thick,->,lightgray] (A312) edge (X312);        
	\path[thick,->,lightgray] (A313) edge (X313);        

	\node[annot] at (\la,-0.3) {$X^0_{s\star}$};	
	\node[annot] at (\lb,-0.3) {$A^1_{s\star}$};	
	\node[annot] at (\lc,-0.3) {$X^1_{s\star}$};	
	\node[annot] at (\ld,-0.3) {$A^2_{s\star}$};	
	\node[annot] at (\le,-0.3) {$X^2_{s\star}$};	
	\node[annot] at (\lf,-0.3) {$A^3_{s\star}$};	
	\node[annot] at (\lg,-0.3) {$X^3_{s\star}$};	

	\node[annot] at ({(\la+\lb)*0.5},-0.1) {$W^1$};
	\node[annot] at ({(\lc+\ld)*0.5},-0.1) {$W^2$};
	\node[annot] at ({(\le+\lf)*0.5},-0.1) {$W^3$};
	
	\def\varia{0.3}
	\fill[scale=0.5,domain=-1.2:1.2,smooth,variable=\x,lightgray,shift={(4.0,6.0)},samples=100] plot (\x,{1.3*exp(-(\x)^2*2000))});
	\draw[scale=0.5,domain=-1.2:1.2,smooth,variable=\x,black,shift={(4.0,6.0)},samples=100] plot (\x,{1.3*exp(-(\x)^2*2000))});
	
	\fill[scale=0.5,domain=-1.2:1.2,smooth,variable=\x,lightgray,shift={(6.8,6.0)}] plot (\x,{1/(sqrt(2*pi)*\varia) * exp(-(\x)^2/(2*\varia^2))});
    \draw[scale=0.5,domain=-1.2:1.2,smooth,variable=\x,black,shift={(6.8,6.0)}] plot (\x,{1/(sqrt(2*pi)*\varia) * exp(-(\x)^2/(2*\varia^2))});
    
	\fill[scale=0.5,domain=-1.2:1.2,smooth,variable=\x,lightgray,shift={(10.8,6.0)}] plot (\x,{1/(sqrt(2*pi)*\varia) * exp(-(\x)^2/(2*\varia^2))});    
    \draw[scale=0.5,domain=-1.2:1.2,smooth,variable=\x,black,shift={(10.8,6.0)}] plot (\x,{1/(sqrt(2*pi)*\varia) * exp(-(\x)^2/(2*\varia^2))});
    
    \fill[scale=0.5,domain=-1.2:1.2,smooth,variable=\x,lightgray,shift={(13.6,6.0)}] plot (\x,{0.5/(sqrt(2*pi)*\varia) * exp(-(\x+0.4)^2/(2*\varia^2)) + 0.5 * 1/(sqrt(2*pi)*\varia) * exp(-(\x-0.2)^6/(2*\varia^2))});
    \draw[scale=0.5,domain=-1.2:1.2,smooth,variable=\x,black,shift={(13.6,6.0)}] plot (\x,{0.5/(sqrt(2*pi)*\varia) * exp(-(\x+0.4)^2/(2*\varia^2)) + 0.5 * 1/(sqrt(2*pi)*\varia) * exp(-(\x-0.2)^6/(2*\varia^2))});
    
	\fill[scale=0.5,domain=-1.2:1.2,smooth,variable=\x,lightgray,shift={(17.5,6.0)}] plot (\x,{1/(sqrt(2*pi)*\varia) * exp(-(\x)^2/(2*\varia^2))});        
	\draw[scale=0.5,domain=-1.2:1.2,smooth,variable=\x,black,shift={(17.5,6.0)}] plot (\x,{1/(sqrt(2*pi)*\varia) * exp(-(\x)^2/(2*\varia^2))});    
	
\end{tikzpicture}
\caption{
The principle of the normal approximation in a feed-forward \GPN{} network.
Propagating the distribution of the activations $A^2_{s\star}$ through the activation functions results in arbitrary distributions $X^2_{s\star}$.
However, given enough \GPNs{} in layer 1 and 2, the responses $X^2_{sn}$, $n \in \{1, 2, \dots N_2\}$, are only weakly correlated and thus the distribution of the activations $A^3_{s\star}$ will again resemble a normal distribution due to the central limit theorem.
Other random variables (virtual observations, responses) are omitted from this graph.
}
\label{fig:gpn_parametric_normal_approx}
\end{figure}

Consider the excerpt of a \GPN{} feed-forward network shown in \cref{fig:gpn_parametric_normal_approx}.
Since the inputs are observed, the distribution of each \GPN{} in layer 1 is a univariate normal distribution with all \GPNs{} being pairwise independent.
The activations of the \GPNs{} in layer 2 are given by a matrix multiplication of the outputs of layer 1, thus they follow a multivariate normal distribution.
In general the distribution of the outputs of the \GPNs{} in layer 2 is not normal, since the activation function can transform the incoming distribution arbitrarily.

The central limit theorem for weakly dependent random variables~\cite{Lehmann200408} states that if $Z_1, \dots, Z_N$ are $N$ random variables with $\E [Z_i] = \mu$ and $\Var (Z_i) = \sigma^2$, then 
\begin{equation}\label{eq:clp_dep_var}
S_N \teq \sqrt{N} \left[ \left( \frac{1}{N} \sum_{n=1}^N Z_n \right) - \mu \right] 
\end{equation}
converges to a normal distribution in the limit $N \to \infty$, provided that
\begin{equation}
\lim_{N \to \infty} \frac{1}{N} \sum_{i=1}^N \sum_{\substack{j=1 \\ j \neq i}}^N \Cov(Z_i, Z_j) < \infty \,.
\end{equation}
For the activations of a \GPN{} with index $n$ within layer 3 this condition evaluates to
\begin{equation}\label{eq:clt_dep_tau}
\lim_{N_2 \to \infty} \frac{1}{N_2} \sum_{m=1}^{N_2} \sum_{\substack{m'=1 \\ m \neq m'}}^{N_2} \Cov\!\left( X^2_{sm} W^3_{mn}, X^2_{sm'} W^3_{m'n} \right) < \infty \,.
\end{equation}
If the weights $W^3$ are initialized randomly from an \iid{} distribution, it is clear that this condition holds before training is performed and thus $A^3_{s\star}$ converges to a normal distribution.
However, it is not possible to predict how the weight distribution develops during training.
Thus it is necessary to turn to an empirical analysis of their behavior.
\cite{Wang:2013} performed such an analysis on the activations of a conventional neural network using a sigmoidal activation function.
They showed empirically that the activation of each neuron remains normally distributed even \emph{after} the network has been trained to convergence, provided that the number of incoming connections is large enough.

\begin{figure}[tb]\centering
\begin{subfigure}{0.37\textwidth}
\centering
\includegraphics[width=0.99\columnwidth]{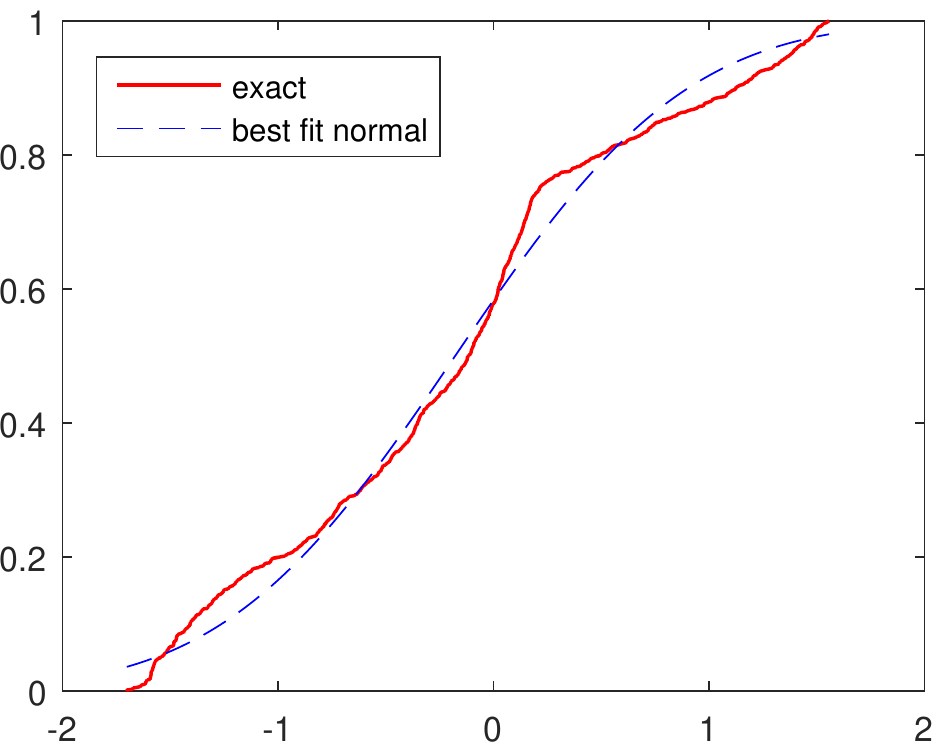}
\caption{layer 1 and 2 both have 3 \GPNs{}}
\end{subfigure}\hspace{9mm}
\begin{subfigure}{0.37\textwidth}
\centering
\includegraphics[width=0.99\columnwidth]{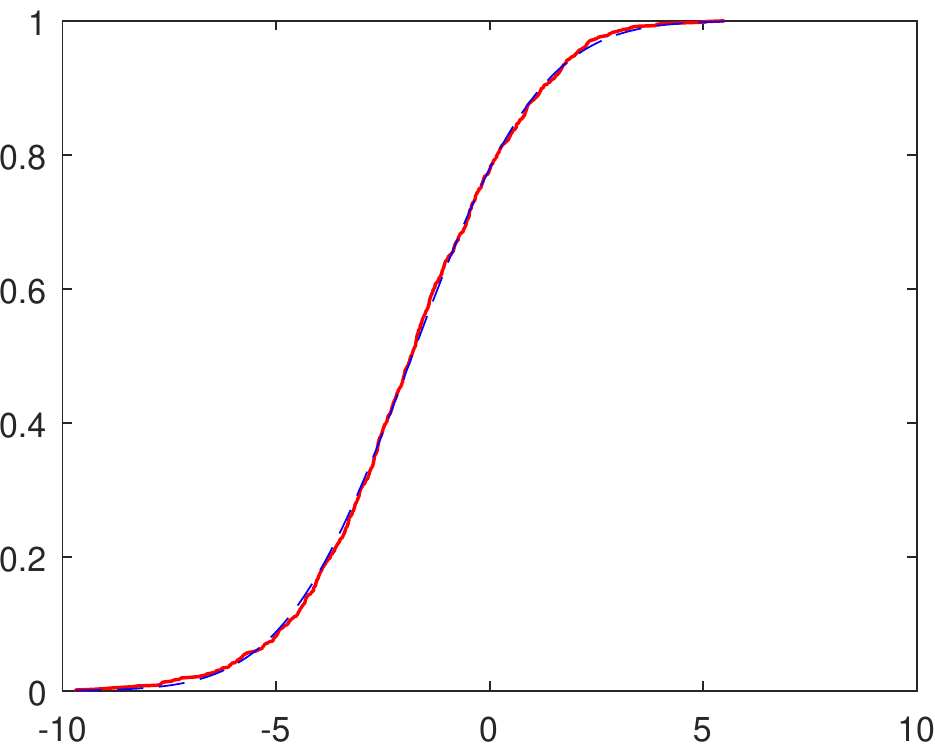}
\caption{layer 1 and 2 both have 10 \GPNs{}}
\end{subfigure}
\caption{
The figure shows the \CDF{} of the distribution of activations of a \GPN{} in layer~3 of the \GPN{} feed-forward network shown in \cref{fig:gpn_parametric_normal_approx}.
The red line shows the empirical distribution obtained by propagating 1000 draws from $X^1_{s\star}$ through layer~2 and applying the weights $W^3$.
The dashed blue line is the \CDF{} of a best-fit normal distribution on this data.
}
\label{fig:gpn_parametric_normal_approx_cdf}
\end{figure}

To determine how many \GPNs{} are necessary to obtain a reasonable approximation of the normal distribution, we perform the following experiment.
We instantiate a three-layer \GPN{} feed-forward network, as in \cref{fig:gpn_parametric_normal_approx}, with different numbers of \GPNs{} in layer 1 and 2.
Their weights are sampled from a standard normal distribution and the activation function of each \GPN{} is randomly sampled from a zero-mean \Gp{} with the \Se{} covariance function with unit lengthscale.
A random input vector is drawn and the parameters of the distribution for $X^1_{s\star}$ are calculated.
Then 1000 samples are drawn from $X^1_{s\star}$ and propagated through each network until reaching $A^3_{s\star}$.
The resulting empirical \acp{CDF} of $A^3_{s\star}$ for two networks are shown in \cref{fig:gpn_parametric_normal_approx_cdf} together with the best-fit normal \acp{CDF}.
For a very small network with only 3~\GPNs{} on both layers it is apparent that the activations are not normally distributed.
As the number of \GPNs{} increases the empirical distribution approaches a normal distribution.
Having 10~\GPNs{} in both layers is enough for the distribution to resemble a Gaussian very closely.

We have thus motivated analytically and verified empirically that assuming a multivariate normal distribution over the \emph{activations} $A^l_{s\star}$ for all layers $l$ is a reasonable choice for non-degenerate networks of parametric \GPNs{}.
The parameters for this multivariate normal, 
\begin{equation}
\Ptilde(A^{l}_{s\star}) = \N (A^{l}_{s\star} \| \mutilde^{A^{l}}_{s\star}, \Sigmatilde^{A^{l}}_{s\star\star} )  \,,
\end{equation} 
are obtained by calculating the expectation and covariance,
\begin{subequations}
\begin{align}
\mutilde^{A^{l}}_{sn} &\teq \E_{\P(A^{l}_{s\star} \| A^{l-1}_{s\star}) \Ptilde(A^{l-1}_{s\star}) } [A^{l}_{sn}] \,, \\
\Sigmatilde^{A^{l}}_{snm} &\teq \Cov_{\P(A^{l}_{s\star} \| A^{l-1}_{s\star}) \Ptilde(A^{l-1}_{s\star})} (A^{l}_{sn}, A^{l}_{sm}) \,.
\end{align}
\end{subequations}
It turns out that both can be evaluated \emph{exactly} and in closed form as we will show in the following section.
Note that here $\Sigmatilde$ describes the covariance between different \GPNs{} within a layer while $\Sigma$ from~\eqref{eq:gpn_parametrized_sigma_y_sn} captures the covariance between different samples for the same parametric \GPN{}.

\subsection{Propagation of Mean and Covariance}\label{sec:gpn_propagation}
Since $A^l_{s\star} = X^{l-1}_{s\star} \, W^l$ we have for each layer
\begin{equation}\label{eq:gpn_normal_a} 
\mutilde^{A^l}_{s\star} = \mutilde^{X^{l-1}}_{s\star} \,  W^l \,, \quad\quad\quad
\Sigmatilde^{A^l}_{s\star\star} = (W^l)^T \, \Sigmatilde^{X^{l-1}}_{s\star\star} \, W^l \,, 
\end{equation}
where $\mutilde^{X^{l-1}}$ and $\Sigmatilde^{X^{l-1}}$ are the mean and covariance of $X^{l-1}$, which is not necessarily normally distributed.
We now have to calculate $\mutilde^{X^{l}}$ and $\Sigmatilde^{X^{l}}$ given $\mutilde^{A^l}$ and $\Sigmatilde^{A^l}$.
Note that we do \emph{not} imply that $X^l$ is normally distributed, we just use the symbols $\mutilde^{X^{l}}$ and $\Sigmatilde^{X^{l}}$ to denote the mean and covariance of it.

The first layer, $l=1$, already follows a normal distribution, since its inputs and thus activations are deterministic.
Thus we set
\begin{equation}
\mutilde^{X^1}_{sn}     = \mu^{X^1}_{sn} \,, \quad\quad\quad
\Sigmatilde^{X^1}_{snm} = \Sigma^{X^1}_{ssn} \, \delta_{nm} \,
\end{equation}
with $\mu^{X^1}_{sn}$ and $\Sigma^{X^1}_{ssn}$ given by~\eqref{eq:gpn_parametrized_mu_sigma_y_sn} and $\delta_{nm}$ being the Kronecker delta.

For the following layers, $l > 1$, we proceed as follows.
To calculate the mean vector we apply the law of total expectation and get
\begin{equation}
\mutilde^{X^{l}}_{sn} = \E [X^{l}_{sn}] =
\E_{\Ptilde(A^{l}_{s\star})} \left[ \E_{\P(X^{l}_{s\star} \| A^{l}_{s\star})} [ X^{l}_{sn} ] \right]
= \E_{\Ptilde(A^{l}_{s\star})} [ \mu^{X^{l}}_{sn} ] \,.
\end{equation}
By inserting $\mu^{X^{l}}_{sn}$ from~\eqref{eq:gpn_parametrized_mu_y_sn} we further obtain
\begin{equation}\label{eq:gpn_normal_mu}
\mutilde^{X^{l}}_{sn} = \left( \E_{\Ptilde(A^{l}_{s\star})} [ \alpha^l_{s \star n} ] \right)^T \kappa^l_{\star\star n} \,  U^l_{\star n}
\end{equation}
with
\begin{equation}\label{eq:gpn_normal_alpha_kappa} 
\alpha^l_{srn} \teq k^{\lambda^l_n}_{\SE{}}(A^l_{sn}, V^l_{rn})  \,, \quad\quad\quad
\kappa^l_{\star\star n} \teq [K^l_n(V^l_{\star n}, V^l_{\star n}) + \diag (S^l_{\star n})]^{-1}  \,. 
\end{equation}
Noting that the \Se{} covariance function can be written as the unnormalized \PDF{} of a Gaussian and applying the product formula for two Gaussian \PDFs{} gives~\cite{bromiley2003products,quinonero2002prediction}
\begin{equation}\label{eq:gpn_normal_E_alpha}
\psi^l_{rn} \teq \E [ \alpha^l_{rn} ] = \sqrt{\frac{(\lambda^l_n)^2}{(\lambda^l_n)^2 + \Sigmatilde^{A^l}_{snn}}} \, \exp\!\left( -\frac{(\mutilde^{A^l}_{sn}  - V^l_{rn})^2}{2 ((\lambda^l_n)^2 + \Sigmatilde^{A^l}_{snn})} \right) \,.
\end{equation}
This concludes the calculation of the mean of $X^l$.

For the covariance matrix we obtain by expanding the occurring expectations 
\begin{equation}
\Sigmatilde^{X^{l}}_{snm} = \Cov (X^{l}_{sn}, X^{l}_{sm}) =  
\E_{\Ptilde(A^{l}_{s\star})}\!\left[ \E_{\P(X^{l}_{s\star} \| A^{l}_{s\star})} [ X^{l}_{sn} \, X^{l}_{sm} ] \right] - \mutilde^{X^{l}}_{sn} \, \mutilde^{X^{l}}_{sm} \,.
\end{equation}
We must now differentiate between on-diagonal and off-diagonal elements of the covariance matrix.
For elements representing the variance, $n=m$, we note that $\E[X^2] = \Var(X) + \E[X]^2$ and obtain   
\begin{equation}\label{eq:gpn_normal_Sigmatilde_1}
\Sigmatilde^{X^{l}}_{snn} = \E_{\Ptilde(A^{l}_{s\star})}\!\left[ \Sigma^{X^l}_{ssn} + (\mu^{X^l}_{sn})^2 \right] - (\mutilde^{X^l}_{sn})^2  \,,
\end{equation}
with
\begin{align}
\E\!\left[ \Sigma^{X^l}_{ssn} \right] &= 
1 - \E\!\left[ (\alpha^l_{s\star})^T \, \kappa^l_{\star\star n} \,  \alpha^l_{s\star} \right] + (\sigma^l_n)^2  = 1  - \sum_r \sum_t \kappa^l_{rtn} \, \Omega^l_{srtn} + (\sigma^l_n)^2 \,, \label{eq:gpn_normal_sigma_xl} \\
\E\!\left[ (\mu^{X^l}_{sn})^2 \right] &= \E\!\left[ (\beta^l_{\star n})^T \, \alpha^l_{\star n} \, (\alpha^l_{\star n})^T \, \beta^l_{\star n} \right] = \sum_r \sum_t \beta^l_{rn} \beta^l_{tn} \, \Omega^l_{srtn} \label{eq:e_mu_xl_sn}
\end{align}
where $\beta^l_{\star n} \teq \kappa^l_{\star\star n} \, U^l_{\star n}$.
Inserting the derived terms into~\eqref{eq:gpn_normal_Sigmatilde_1} and simplifying gives
\begin{equation}\label{eq:gpn_normal_Sigmatilde}
\Sigmatilde^{X^{l}}_{snn} = 1 - \tr\!\left[ \left( \kappa^l_{\star \star n} - \beta^l_{\star n} \, (\beta^l_{\star n})^T \right)  \Omega^l_{s\star \star n} \right] - \tr\!\left( \psi^l_{\star n} (\psi^l_{\star n})^T \, \beta^l_{\star n} (\beta^l_{\star n})^T \right)  + (\sigma^l_n)^2 \,.
\end{equation}
where the product formula for Gaussian \PDFs{} is used to evaluate $\Omega$ to
\begin{equation}\label{eq:gpn_normal_omega}
\Omega^l_{srtn} \teq \E[\alpha^l_{rn} \, \alpha^l_{tn}] = \sqrt{\frac{(\lambda^l_n)^2}{(\lambda^l_n)^2 + 2 \Sigmatilde^{A^l}_{snn}}} \, \exp\!\left( -\frac{\left(\mutilde^{A^l}_{sn} - \frac{V^l_{rn} + V^l_{tn}}{2} \right)^2}{(\lambda^l_n)^2 + 2 \Sigmatilde^{A^l}_{snn}}  - \frac{(V^l_{rn} - V^l_{tn})^2}{4 (\lambda^l_n)^2} \right) \,.
\end{equation}

For off-diagonal elements of the covariance matrix, $n \neq m$, we observe that $X^{l}_{sn}$ and $X^{l}_{sm}$ are \emph{conditionally independent} given $A^{l}_{s\star}$ because the activation functions of \GPNs{} $n$ and $m$ are represented by two different \Gps{}.
Exploiting this gives
\begin{align}
\Sigmatilde^{X^l}_{snm} 
&= \E\!\left[ \mu^{X^l}_{sn} \, \mu^{X^l}_{sm} \right] - \mutilde^{X^l}_{sn} \, \mutilde^{X^l}_{sm} 
= \E\!\left[ (\beta^l_{\star n})^T \, \alpha^l_{\star n} \, (\alpha^l_{\star m})^T \, \beta^l_{\star m} \right] - \mutilde^{X^l}_{sn} \, \mutilde^{X^l}_{sm} \nonumber \\
&= \sum_r \sum_t \beta^l_{rn} \beta^l_{tm} \, \Lambda^l_{srtnm} - \mutilde^{X^l}_{sn} \, \mutilde^{X^l}_{sm} \label{eq:gpn_normal_Sigmatilde_cov} 
\end{align}
where the product formula for two-dimensional Gaussian \PDFs{} is used to evaluate $\Lambda$ to
\begin{align}
\Lambda^l_{srtnm} &\teq \E[\alpha^l_{rn} \alpha^l_{tm}] =  \frac{ \lambda^l_n \, \lambda^l_m \, \exp(\mathscr{A} / \mathscr{B} ) }{\sqrt{ [ (\lambda^l_n)^2 + \widetilde{\Sigma}^{A^l}_{snn} ] \, [ (\lambda^l_m)^2 + \widetilde{\Sigma}^{A^l}_{smm} ] - (\widetilde{\Sigma}^{A^l}_{snm})^2}} \,, \label{eq:gpn_normal_lambda} \\
\mathscr{A} &\teq  (V^l_{rm} - \mutilde^{A^l}_{sm})^2 \, [ (\lambda^l_n)^2 + \widetilde{\Sigma}^{A^l}_{snn} ]  +  (V^l_{tn} - \mutilde^{A^l}_{sn})^2 \, [ (\lambda^l_m)^2 + \widetilde{\Sigma}^{A^l}_{smm} ]  \, + \nonumber \\
& \quad\quad  2 \, (V^l_{td} - \mutilde^{A^l}_{sn}) \, (\mutilde^{A^l}_{sm} - V^l_{rm}) \, \widetilde{\Sigma}^{A^l}_{snm}  \,, \nonumber \\
\mathscr{B} &\teq 2 \left\{ \left[ (\lambda^l_n)^2 + \widetilde{\Sigma}^{A^l}_{snn} \right] \, \left[ (\lambda^l_m)^2 + \widetilde{\Sigma}^{A^l}_{smm} \right] -  (\widetilde{\Sigma}^{A^l}_{smn})^2 \right\} \,. \nonumber
\end{align}
This concludes the calculation of the covariance matrix of $X^l$.

We are now able to analytically propagate the mean and covariance of all data points through all layers of a feed-forward \GPN{} stack.
The moments of the distribution $\Ptilde_\theta(X^L_{s\star} \| X^0_{s\star})$ are calculated by iterative applications of \cref{eq:gpn_normal_a,eq:gpn_normal_mu,eq:gpn_normal_Sigmatilde,eq:gpn_normal_Sigmatilde_cov} for $l \in \{1, 2, \dots, L\}$.

\subsection{Evaluating the Loss}\label{sec:eval_loss}
For a regression task, we use the loss measure
\begin{equation}\label{eq:gpn_loss_reg}
L_{\mathrm{reg}} (\vec{y}, \vec{t}) \teq -\delta (\vec{y} - \vec{t}) \,.
\end{equation}
where $\delta(\vec{z})$ is the delta distribution.
Inserting this loss measure into~\eqref{eq:gpn_parametrized_loss} and expanding the expectation over the responses $F^L$ of the last layer leads to
\begin{align}
\mathscr{L}_{\mathrm{reg}}(\theta) 
&= - \frac{1}{S} \sum_{s=1}^S \iint \P(F^{L}_{s\star} \| X^0_{s\star}) \, \P(X^{L}_{s\star} \| F^{L}_{s\star}) \, \delta(X^L_{s\star} - T_{s\star}) \, \d F^{L}_{s\star} \, \d X^L_{s\star} \nonumber \\
&= - \frac{1}{S} \sum_{s=1}^S \E_{\P(F^L_{s\star} \| X^0_{s\star})}\!\left[ \P(X^{L}_{s\star} = T_{s\star}\| F^{L}_{s\star} )  \right] \,. \label{eq:gpn_parametric_regression_loss}
\end{align}
We can further take the logarithm and apply Jensen's inequality to obtain the negative log-likelihood objective function,
\begin{equation}\label{eq:gpn_parametric_regression_log_loss}
- \log (-\mathscr{L}_{\mathrm{reg}}(\theta)) \leq \mathscr{L}_{\mathrm{ll}}(\theta) = - \frac{1}{S} \sum_{s=1}^S \E_{\P(F^{L}_{s\star} \| X^0_{s\star})}\!\left[\log \P(X^{L}_{s\star} = T_{s\star}\| F^{L}_{s\star} ) \right]  \,,
\end{equation}
which is an upper bound of the logarithmic regression loss.
We replace $\P(F^{L}_{s\star} \| X^0_{s\star})$ by the distribution $\Ptilde(F^{L}_{s\star} \| X^0_{s\star})$  obtained by propagating moments as done in \cref{sec:gpn_propagation}.
Evaluating the expectation in~\eqref{eq:gpn_parametric_regression_log_loss} gives
\begin{equation}\label{eq:gpn_reg_loss_marginal_final}
\mathscr{L}_{\mathrm{ll}}(\theta) \propto - S \sum_{n=1}^{N_L} \log \sigma_n^L - \frac{1}{2} \sum_{s=1}^S \sum_{n=1}^{N_L} \frac{(T_{sn})^2 - 2\,T_{sn}\,\E\!\left[ F^L_{sn} \right] + \E\!\left[ (F^L_{sn})^2 \right]}{(\sigma^L_n)^2} \,
\end{equation}
with $\E\!\left[ F^L_{sn} \right] = \mutilde^{X^L}_{sn}$ given by~\eqref{eq:gpn_normal_mu} and 
\begin{equation}
\E\!\left[ (F^L_{sn})^2 \right] = 1 - \tr\!\left[ \left(\kappa^L_{\star\star n} - \beta^L_{\star n} (\beta^L_{\star n})^T \right) \Omega^L_{s \star\star n} \right] \,,
\end{equation}
which was calculated in the same way as~\eqref{eq:gpn_normal_Sigmatilde} and uses $\kappa$, $\beta$, $\Omega$ from that equation.
Thus we obtained a fully deterministic expression for an upper bound of the logarithmic regression loss.

For a classification task we use the cross-entropy loss measure, resulting in the loss
\begin{equation}\label{eq:gpn_parametrized_loss_classification}
\mathscr{L}_{\mathrm{class}}(\theta) \teq \frac{1}{S} \sum_{s=1}^S \E_{\P(X^L_{s\star} \| X^0_{s\star})} \!\left[ T_{s\star} \cdot \log \mathrm{softmax}( A^{L+1}_{s\star} ) \right] \,
\end{equation}  
where $A^{L+1}_{s\star} \teq X^L_{s\star} \, W^{L+1}$ is a final linear combination of the model outputs using the weights $W^{L+1}$.
The targets $T_{s\star}$ use a one-hot encoding for the classes, $\cdot$ denotes the scalar product between two vectors and 
\begin{equation}\label{eq:nn_softmax}
\mathrm{softmax}(\vec{o})_i \teq \frac{\exp o_i}{\sum_j \exp o_j} \,.
\end{equation}
It is not possible to exactly evaluate the expectation in~\eqref{eq:gpn_parametrized_loss_classification} over the softmax function.

A method for approximate evaluation of such expectations without the necessity to sample is the unscented transform~\cite{julier1996general}.
It works by propagating deterministically chosen points that represent the normal distribution of $A^{L+1}_{s\star}$ through the function and using the transformed points to estimate the mean and covariance of the transformed distribution.
Applying the unscented transform the loss becomes
\begin{equation}\label{eq:gpn_normal_unscented}
\mathscr{L}_{\mathrm{class}}(\theta) = \frac{1}{S} \sum_{s=1}^S \sum_{i=0}^{2 N_L} \mathscr{W}_i \, T_{s\star} \cdot \log \mathrm{softmax}( \vec{x}_{si} )
\end{equation}
where $\vec{x}_{si}$ are the $2 N_L + 1$ sigma points with associated weights $\mathscr{W}_i$ given by 
\begin{subequations}\label{eq:unscented_sigma}
\begin{align}
\vec{x}_{s0} 		&\teq \mutilde^{X^L}_{s\star} \, W^{L+1} 			\,, & \mathscr{W}_0     &\teq \frac{\kappa}{N_L + \kappa}  \,, \\
\vec{x}_{si} 		&\teq \mutilde^{X^L}_{s\star} \, W^{L+1} + \mathcal{S}_{i\star}	\,, & \mathscr{W}_i     &\teq \frac{1}{2 \, (N_L + \kappa)} \,, \,\,\, i \in \{1, 2,\dots, N_L\} \,, \\
\vec{x}_{s,i+d} 	&\teq \mutilde^{X^L}_{s\star} \, W^{L+1} - \mathcal{S}_{i\star}   \,,  & \mathscr{W}_{i+d} &\teq \frac{1}{2 \, (N_L + \kappa)}  \,, \,\,\, i \in \{1, 2,\dots, N_L\} \,,
\end{align}
\end{subequations}
where for the parameter $\kappa$ we usually use $\kappa = 3-N_L$ and
\begin{equation}\label{eq:unscented_s}
\mathcal{S} \teq \chol [ (N_L+\kappa) \, (W^{L+1})^T \, \Sigmatilde^{X^{L}}_{s\star\star} \, W^{L+1} ]
\end{equation}
is the Cholesky decomposition of the scaled covariance matrix.
The Cholesky decomposition is differentiable~\cite{smith1995differentiation,murray2016differentiation} and thus the derivative of the loss can be calculated through this operation.
Note that if the covariance matrix $\Sigmatilde^{X^{L}}_{s\star\star}$ is assumed to be diagonal, the Cholesky decomposition is given by the square root of the diagonal elements and is thus inexpensive to compute.

In conclusion, we have derived completely deterministic training objectives for a feed-forward parametric \GPN{} network using analytic propagation of means and covariances from layer to layer, resulting in loss functions that can be minimized using mini-batch gradient descent with derivatives computed using automatic differentiation.
By minimizing the derived losses, the predicted variance of the model predictions $X^{L}$ is taken into account.
A prediction that is far off from the ground truth will be penalized stronger if the \GPN{} stack simultaneously predicts a low variance, \ie high confidence, at the same time.
Consequently during training the model not only learns to predict the targets but also to self-estimate the confidence of its predictions.

\subsection{Computational and Model Complexity}\label{sec:gpn_complexity}
To represent the activation functions a parametric \GPN{} layer requires $3R$ parameters per \GPN{}, where $R$ is the number of virtual observations. 
\GPNs{} require no bias term, because it is equivalent to an offset in the targets corresponding to the inducing points.
To reduce the number of parameters the locations of the inducing points $V$ can be fixed, for example using equidistantly placed points, so that only $2R$ parameters are required.
Furthermore, the standard deviations $S$ of the targets $U$ can be shared between all inducing points, resulting in only $R+1$ parameters per \GPN{}.
Finally, to further reduce the number of required parameters we can apply the weight sharing idea from \CNNs{} and use a common set of virtual observations and thus activation functions within a group of \GPNs{} or even within a whole layer.

\begin{figure}
\centering
\begin{subfigure}{0.4\columnwidth}
\centering
\includegraphics[width=0.99\columnwidth]{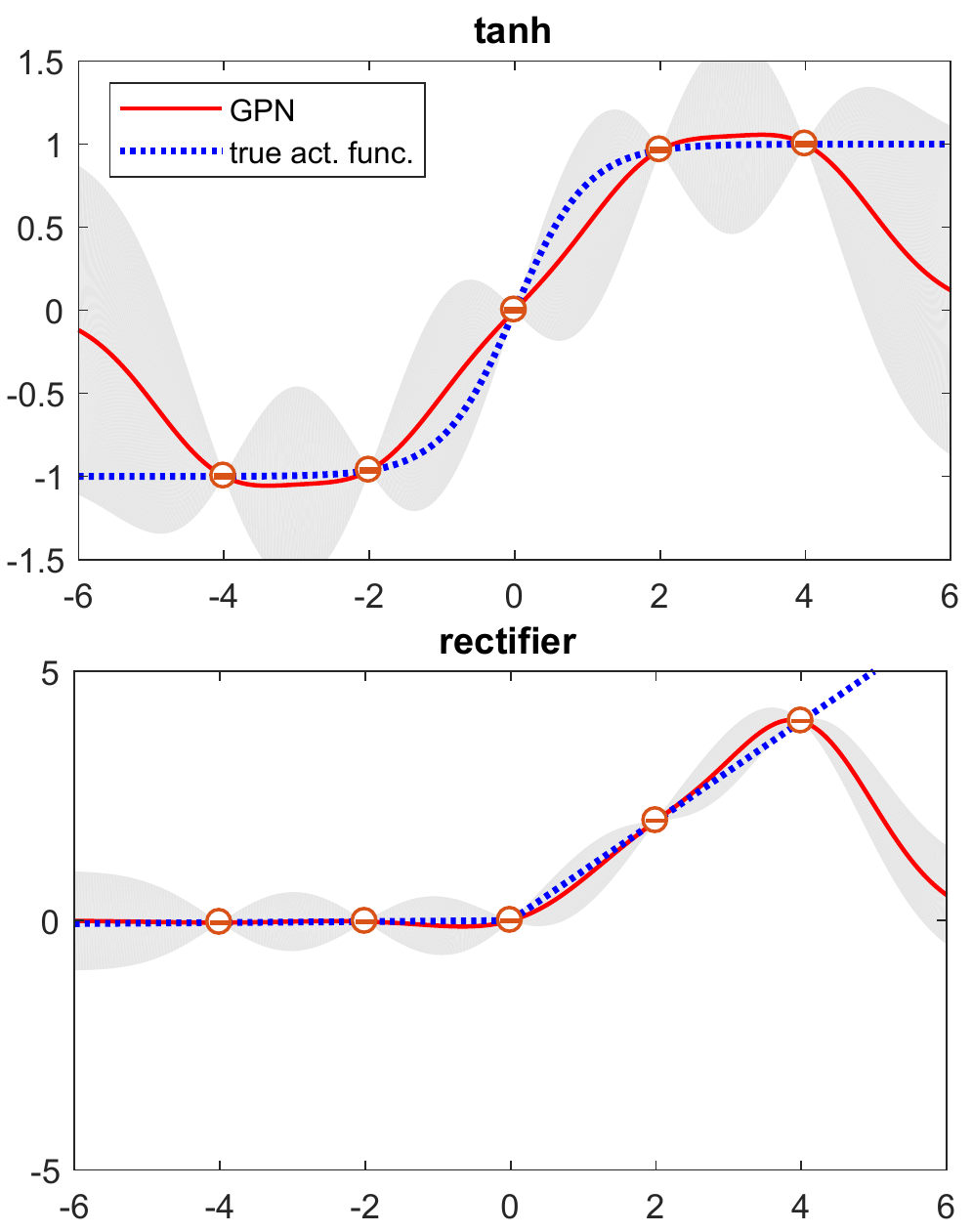}
\caption{\GPN{} using 5 virtual observations}
\end{subfigure} 
\begin{subfigure}{0.4\columnwidth}
\centering
\includegraphics[width=0.99\columnwidth]{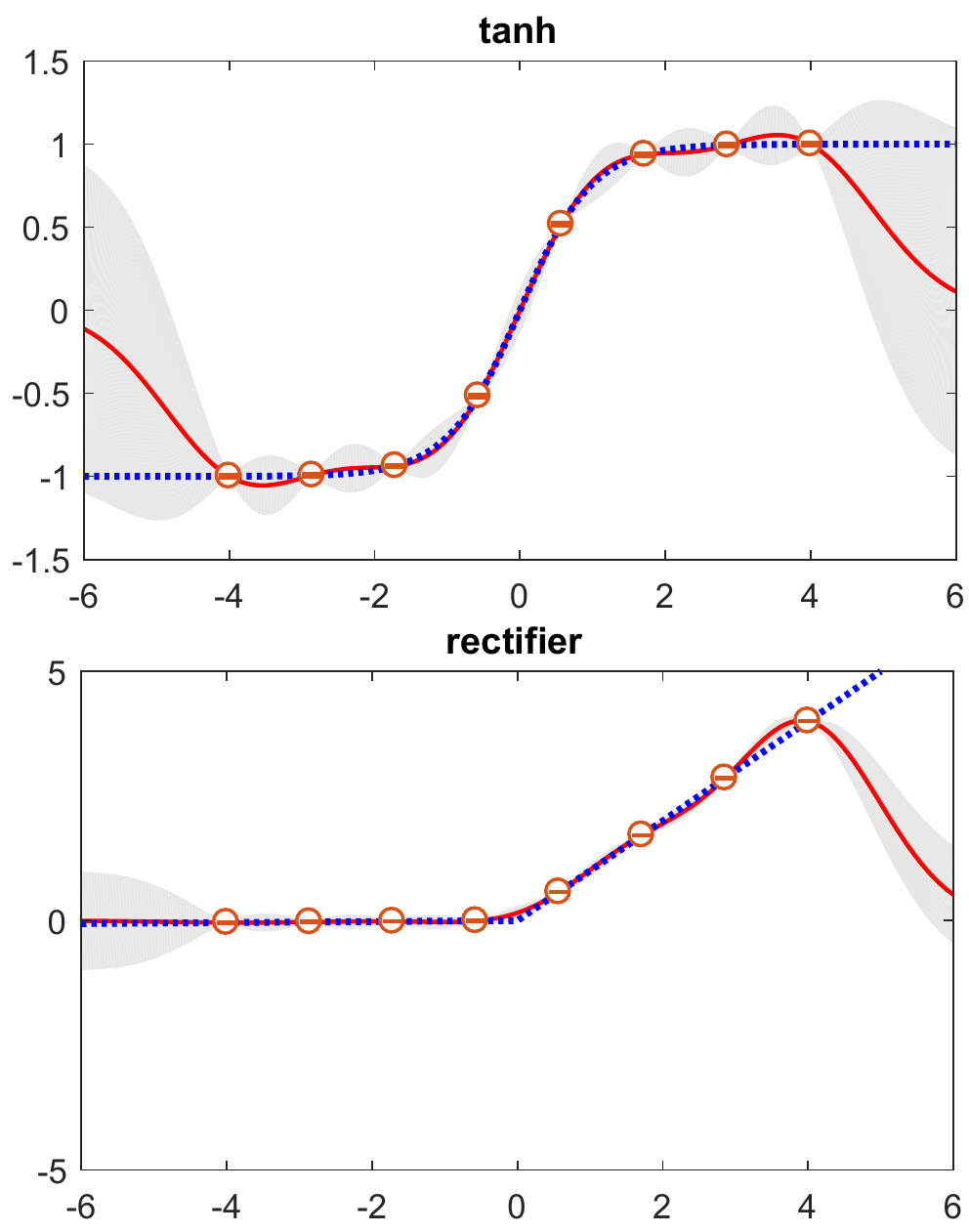}
\caption{\GPN{} using 8 virtual observations}
\end{subfigure}
\caption{
Common activation functions approximated by a parametric \GPN{} with 5 or 8 virtual observations (red circles) respectively. 
The dotted line shows the actual activation functions, while the red line is the approximation by the parametric \GPN{}.
The standard deviation is shown using gray shading.
}
\label{fig:gpn_standards_act_fns}
\end{figure}

Since the flexibility of the parametric \GPN{} is controlled by $R$, care must be taken not to choose a too small $R$ since this would limit the activation functions representable by the \GPN{} and thus the power of a feed-forward network constructed from these \GPNs{}.
A good heuristic for choosing $R$ is so that the \GPN{} is able to represent the most common activation functions currently in use.
For that purpose, we empirically modeled the hyperbolic tangent and rectifier activation functions with a \GPN{} using a varying number of inducing points and compared the resulting approximation to the original function.
From \cref{fig:gpn_standards_act_fns} we can see that $R=8$ virtual observations with equidistant inducing points are enough to represent these activation functions with high accuracy.

\begin{table}[bt]
\centering
\begin{tabular}{lll|rr} \toprule
\emph{expression} &&& \multicolumn{2}{c}{\emph{complexity}} \\
\hline
\multicolumn{3}{c|}{\emph{activations given} $\mutilde^{X^{l-1}}_{s\star}$, $\Sigmatilde^{X^{l-1}}_{s\star}$} && \\
mean & $\mutilde^{A^l}_{s\star}$ & \eqref{eq:gpn_normal_a} 	& \multicolumn{2}{c}{$\O(N_l \, N_{l-1})$} \\
variance & $\diag(\Sigmatilde^{A^l}_{s\star\star})$ & \eqref{eq:gpn_normal_a} & \multicolumn{2}{c}{$\O(N_l \, N_{l-1})$} \\
covariance & $\Sigmatilde^{A^l}_{s\star\star}$ & \eqref{eq:gpn_normal_a} & \multicolumn{2}{c}{$\O(N_l^2\,N_{l-1} + N_l\,N^2_{l-1})$} \\
\hline
\multicolumn{3}{c|}{\emph{responses given} $\mutilde^{A^{l}}_{s\star}$, $\Sigmatilde^{A^{l}}_{s\star}$} & \emph{variable} $V,S$ & \emph{fixed} $V,S$ \\
mean & $\mutilde^{X^l}_{s\star}$ & \eqref{eq:gpn_normal_mu} & $\O(N_l \, R^{3})$ & $\O(N_l \, R^2)$ \\
variance & $\diag(\Sigmatilde^{X^l}_{s\star\star})$ & \eqref{eq:gpn_normal_Sigmatilde}  & $\O(N_l \, R^{3})$ & $\O(N_l \, R^2)$ \\
covariance & $\Sigmatilde^{X^l}_{s\star\star}$ & \eqref{eq:gpn_normal_Sigmatilde_cov}  & $\O(N_l^2 \, R^{3})$ & $\O(N_l^2 \, R^2)$ \\
\bottomrule
\end{tabular}
\caption{
Computational complexity of propagating the mean and variance or covariance from layer $l-1$ to layer $l$.
The exponent $\O(R^3)$ comes from the complexity of matrix multiplication and inversion.
}
\label{tab:gpn_normal_complexity}
\end{table}

The computational complexities of propagating mean and covariance from layer to layer are shown in \cref{tab:gpn_normal_complexity}.
The complexity of calculating the responses can be significantly reduced from $\O(N_l \, R^{3})$ to $\O(N_l \, R^{2})$ by keeping the inducing points $V^l$ and variances $S^l$ fixed, because in this case the tensor $\kappa^l$ given by~\eqref{eq:gpn_normal_alpha_kappa} is fixed and can be precomputed.
Another method to save computational complexity is to only propagate the variances, \ie diagonal of the covariance matrix, through the \GPN{} stack.
This reduces the complexity of computing the activations from $\O(N_l^2\,N_{l-1} + N_l\,N^2_{l-1})$ to $\O(N_l \, N_{l-1})$ and the complexity of computing the responses from $\O(N_l^2 \, R^{3})$ to $\O(N_l \, R^{3})$.

Note, that the number of parameters and the computational complexity of propagating the means and covariances only depend on the number of virtual observations and parametric \GPNs{}; therefore the memory requirement is \emph{independent} of the number of training samples and the required training time per epoch scales \emph{linearly} with the number of training samples.
Thus, like a conventional neural network, a parametric feed-forward \GPN{}  can inherently be trained on datasets of unlimited size.

\clearpage
\section{Approximate Bayesian Inference}\label{sec:gpn_bayesian}
Introducing learnable activation functions into a neural network increases its expressive power and thus the associated risk of overfitting when maximum likelihood training is performed.
Thus it is sensible to turn to Bayesian inference instead, which gives a posterior distribution of the model parameters instead of a point estimate.
Bayesian inference requires a prior distribution on the parameters.
The original concept of the non-parametric \GPN{} (\cref{sec:gpn}) was to have a \Gp{} prior on the activation function of each unit; hence the natural choice for a prior on the virtual observations of a parametric \GPN{} is such that the \Gp{} prior on the activation function is restored when the virtual observations are marginalized out.
This is indeed possible and the first use of this prior-restoring technique was in finding inducing points for variational sparse \Gp{} regression~\cite{titsias2009variational}.

We now treat the virtual observation targets $U^l_{\star n}$ of each \GPN{} as random variables with the prior distribution
\begin{equation}
\P(U^l_{\star n} \| V^l_{\star n}) = \N\big( U^l_{\star n} \| \vec{0}, K^l_n(V^l_{\star n}, V^l_{\star n}) \big) \label{eq:gpn_vi_puv} \,
\end{equation}
and set the variance to zero, $S^l_{rn} = 0$, since it has become redundant now that the targets are stochastic themselves.
It can be verified that the marginal $\int \P(F^l_{\star n} \| X^{l-1}, U^l_{\star n}) \, \P(U^l_{\star n} \| V^l_{\star n}) \, \d U^l_{\star n}$ has exactly the same distribution as $\P(F^l_{\star n} \| X^{l-1})$ from~\eqref{eq:F_given_X}.

\begin{figure}[tb]
\centering
\raisebox{1mm}{(a)} \begin{tikzpicture}
    \tikzstyle{unit}=[circle,red,very thick,draw,minimum size=14pt,text=black]
	\tikzstyle{annot} = [text width=4em, text centered]	

	\node[unit,label=270:{$X^0$},fill] (X0) at (-0.5,0.0) {};

	\node[unit,label=270:{$A^1$}] (A1) at (1.0,0.0) {};
	\node[unit,label=090:{$U^1$}] (U1) at (2.0,1.0) {};
	\node[unit,label=270:{$F^1$}] (F1) at (2.0,0.0) {};
	\node[unit,label=270:{$X^1$}] (X1) at (3.0,0.0) {};

	\node[unit,label=270:{$A^2$}] (A2) at (4.5,0.0) {};
	\node[unit,label=090:{$U^2$}] (U2) at (5.5,1.0) {};
	\node[unit,label=270:{$F^2$}] (F2) at (5.5,0.0) {};
	\node[unit,label=270:{$X^2$}] (X2) at (6.5,0.0) {};
	
	\node[unit,label=270:{$A^3$}] (A3) at (8.0,0.0) {};
	\node[unit,label=090:{$U^3$}] (U3) at (9.0,1.0) {};
	\node[unit,label=270:{$F^3$}] (F3) at (9.0,0.0) {};
	\node[unit,label=270:{$X^3$},fill] (X3) at (10.0,0.0) {};	

	\path[thick,->] (X0) edge (A1);        
	\path[thick,->] (A1) edge (F1);        
	\path[thick,->] (U1) edge (F1);        
	\path[thick,->] (F1) edge (X1);        

	\path[thick,->] (X1) edge (A2);        
	\path[thick,->] (A2) edge (F2);        
	\path[thick,->] (U2) edge (F2);        
	\path[thick,->] (F2) edge (X2);        

	\path[thick,->] (X2) edge (A3);        
	\path[thick,->] (A3) edge (F3);        
	\path[thick,->] (U3) edge (F3);        
	\path[thick,->] (F3) edge (X3);        

\end{tikzpicture} \\
\vspace{2mm}
\raisebox{1.5mm}{(b)} \begin{tikzpicture}
    \tikzstyle{unit}=[circle,red,very thick,draw,minimum size=14pt,text=black]
	\tikzstyle{annot} = [text width=4em, text centered]	

	\node[unit,label=270:{$X^0$},fill] (X0) at (-0.5,0.0) {};

	\node[unit,label=270:{$A^1$}] (A1) at (1.0,0.0) {};
	\node[unit,label=090:{$U^1$},dotted] (U1) at (2.0,1.0) {};
	\node[unit,label=270:{$F^1$}] (F1) at (2.0,0.0) {};
	\node[unit,label=270:{$X^1$}] (X1) at (3.0,0.0) {};

	\node[unit,label=270:{$A^2$}] (A2) at (4.5,0.0) {};
	\node[unit,label=090:{$U^2$},dotted] (U2) at (5.5,1.0) {};
	\node[unit,label=270:{$F^2$}] (F2) at (5.5,0.0) {};
	\node[unit,label=270:{$X^2$}] (X2) at (6.5,0.0) {};
	
	\node[unit,label=270:{$A^3$}] (A3) at (8.0,0.0) {};
	\node[unit,label=090:{$U^3$},dotted] (U3) at (9.0,1.0) {};
	\node[unit,label=270:{$F^3$}] (F3) at (9.0,0.0) {};
	\node[unit,label=270:{$X^3$},fill] (X3) at (10.0,0.0) {};	

	\path[thick,->] (X0) edge (A1);        
	\path[thick,->] (A1) edge (F1);        
	\path[thick,->] (U1) edge (F1);        
	\path[thick,->] (F1) edge (X1);        

	\path[thick,->] (X1) edge (A2);        
	\path[thick,->] (A2) edge (F2);        
	\path[thick,->] (U2) edge (F2);        
	\path[thick,->] (F2) edge (X2);        

	\path[thick,->] (X2) edge (A3);        
	\path[thick,->] (A3) edge (F3);        
	\path[thick,->] (U3) edge (F3);        
	\path[thick,->] (F3) edge (X3);    

	\def\varia{0.3}
	\fill[scale=0.5,domain=-1.2:1.2,smooth,variable=\x,lightgray,shift={(2.0,0.8)}] plot (\x,{1/(sqrt(2*pi)*\varia) * exp(-(\x)^2/(2*\varia^2))});    
    \draw[scale=0.5,domain=-1.2:1.2,smooth,variable=\x,black,shift={(2.0,0.8)}] plot (\x,{1/(sqrt(2*pi)*\varia) * exp(-(\x)^2/(2*\varia^2))});
    
	\fill[scale=0.5,domain=-1.2:1.2,smooth,variable=\x,lightgray,shift={(9.0,0.8)}] plot (\x,{1/(sqrt(2*pi)*\varia) * exp(-(\x)^2/(2*\varia^2))});    
    \draw[scale=0.5,domain=-1.2:1.2,smooth,variable=\x,black,shift={(9.0,0.8)}] plot (\x,{1/(sqrt(2*pi)*\varia) * exp(-(\x)^2/(2*\varia^2))});

	\fill[scale=0.5,domain=-1.2:1.2,smooth,variable=\x,lightgray,shift={(16.0,0.8)}] plot (\x,{1/(sqrt(2*pi)*\varia) * exp(-(\x)^2/(2*\varia^2))});    
    \draw[scale=0.5,domain=-1.2:1.2,smooth,variable=\x,black,shift={(16.0,0.8)}] plot (\x,{1/(sqrt(2*pi)*\varia) * exp(-(\x)^2/(2*\varia^2))});

    \fill[scale=0.5,domain=-1.2:1.2,smooth,variable=\x,lightgray,shift={(12.8,0.8)}] plot (\x,{0.5/(sqrt(2*pi)*\varia) * exp(-(\x+0.4)^2/(2*\varia^2)) + 0.5 * 1/(sqrt(2*pi)*\varia) * exp(-(\x-0.2)^6/(2*\varia^2))});
    \draw[scale=0.5,domain=-1.2:1.2,smooth,variable=\x,black,shift={(12.8,0.8)}] plot (\x,{0.5/(sqrt(2*pi)*\varia) * exp(-(\x+0.4)^2/(2*\varia^2)) + 0.5 * 1/(sqrt(2*pi)*\varia) * exp(-(\x-0.2)^6/(2*\varia^2))});

    \fill[scale=0.5,domain=-1.2:1.2,smooth,variable=\x,lightgray,shift={(5.8,0.8)}] plot (\x,{0.5/(sqrt(2*pi)*\varia) * exp(-(\x+0.4)^2/(2*\varia^2)) + 0.5 * 1/(sqrt(2*pi)*\varia) * exp(-(\x-0.2)^6/(2*\varia^2))});
    \draw[scale=0.5,domain=-1.2:1.2,smooth,variable=\x,black,shift={(5.8,0.8)}] plot (\x,{0.5/(sqrt(2*pi)*\varia) * exp(-(\x+0.4)^2/(2*\varia^2)) + 0.5 * 1/(sqrt(2*pi)*\varia) * exp(-(\x-0.2)^6/(2*\varia^2))});

\end{tikzpicture} \\
\vspace{2mm}
\raisebox{1mm}{(c)} \begin{tikzpicture}
    \tikzstyle{unit}=[circle,red,very thick,draw,minimum size=14pt,text=black]
	\tikzstyle{annot} = [text width=4em, text centered]	

	\node[unit,label=270:{$X^0$},fill] (X0) at (-0.5,0.0) {};

	\node[unit,label=270:{$A^1$}] (A1) at (1.0,0.0) {};
	\node[unit,label=090:{$U^1$},dotted] (U1) at (2.0,1.0) {};
	\node[unit,label=270:{$F^1$}] (F1) at (2.0,0.0) {};
	\node[unit,label=270:{$X^1$},dotted] (X1) at (3.0,0.0) {};

	\node[unit,label=270:{$A^2$}] (A2) at (4.5,0.0) {};
	\node[unit,label=090:{$U^2$},dotted] (U2) at (5.5,1.0) {};
	\node[unit,label=270:{$F^2$}] (F2) at (5.5,0.0) {};
	\node[unit,label=270:{$X^2$},dotted] (X2) at (6.5,0.0) {};
	
	\node[unit,label=270:{$A^3$}] (A3) at (8.0,0.0) {};
	\node[unit,label=090:{$U^3$},dotted] (U3) at (9.0,1.0) {};
	\node[unit,label=270:{$F^3$}] (F3) at (9.0,0.0) {};
	\node[unit,label=270:{$X^3$},fill] (X3) at (10.0,0.0) {};	

	\path[thick,->] (X0) edge (A1);        
	\path[thick,->] (A1) edge (F1);        
	\path[thick,->] (U1) edge (F1);        

	\path[thick,->] (X1) edge (A2);        
	\path[thick,->] (A2) edge (F2);        
	\path[thick,->] (U2) edge (F2);        
	
	\path[thick,->] (X2) edge (A3);        
	\path[thick,->] (A3) edge (F3);        
	\path[thick,->] (U3) edge (F3);        
	\path[thick,->] (F3) edge (X3);

\end{tikzpicture} 
\caption{
A \GPN{} feed-forward network distribution for three layers and two variational approximations of its posterior.
Each node corresponds to all samples and \GPN{} units within a layer.
Variational parameters are shown as dotted nodes. 
(a) Exact posterior distribution $\P(\{X\}_1^{L-1}, \{A\}_1^L, \{F\}_1^L, \{U\}_1^L \| X^0, X^L)$ that results in a \emph{non-parametric} \GPN{} feed-forward network when marginalized over $\{U\}_1^L$.
(b) Variational approximation of the inducing targets $U^l$ using the central limit distribution for the marginals of the latent activations $A^l$.
(c) Variational mean-field approximation as performed for deep \Gps{} factorizes over the variables $X^l$.
}
\label{fig:gpn_vi_distrs}
\end{figure}
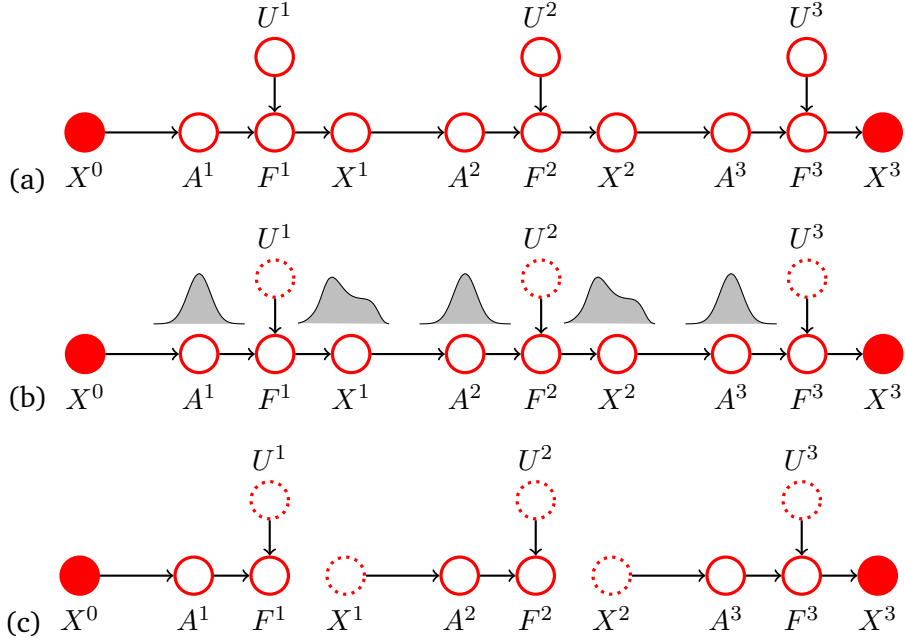

The joint distribution of a parametric \GPN{} feed-forward network is thus given by
\begin{equation}\label{eq:gpn_vi_joint}
\P(\{X\}_1^L, \{A\}_1^L, \{U\}_1^L, \{F\}_1^L \| X^0) = \prod_{l=1}^L \P(A^l \| X^{l-1}) \, \P(U^l) \, \P(F^l \| A^l, U^l) \, \P(X^l \| F^l) \,.
\end{equation}
A graphical model corresponding to that distribution with training targets $X^L$ observed is shown in \cref{fig:gpn_vi_distrs}a.
Because exact inference is intractable we use variational inference~\cite{Bishop:2006} to compute an approximate Bayesian posterior of the model parameters.

Since the information about the activation functions learned from the training data is mediated via the virtual observation targets $U^l$, their variational posterior must be adaptable in order to store that information.
Hence, we choose a normal distribution factorized over the \GPN{} units within a layer with free mean and covariance for the variational posterior of $U^l$,
\begin{align}
\Q(U^l) \teq \prod_{n=1}^{N_l} \Q(U^l_{\star n}) \,,
\quad\quad\quad
\Q(U^l_{\star n}) \teq \N(U^l_{\star n} \| \muhat^{U^l}_{\star n}, \Sigmahat^{U^l}_{\star\star n}) \,.
\end{align}
This allows the inducing targets of a \GPN{} to be correlated, but the covariance matrix can be constrained to be diagonal, if it is desired to reduce the number of model parameters.
We keep the rest of the model distribution unchanged from the prior.
Thus the overall variational posterior is given by
\begin{equation}\label{eq:gpn_vi_stochastic_q}
\Q(\{U\}_1^L, \{X\}_1^{L-1}, \{A\}_1^L, \{F\}_1^L) \teq \prod_{l=1}^L \P(A^l \| X^{l-1}) \, \Q(U^l) \, \P(F^l \| A^l, U^l) \, \P(X^l \| F^l) \,,
\end{equation}
We further assume that each \emph{marginal} $\Q(A^l)$ for $l \in \{1,2, \dots, L\}$ is a normal distribution with appropriate mean and covariance, \ie
\begin{equation}\label{eq:gpn_vi_qal}
\Q(A^l) = \prod_{s=1}^S \Q(A^l_{s\star})  \,, 
\quad\quad\quad
\Q(A^l_{s\star}) = \N\left( A^l_{s\star} \,\Big|\, \mutilde^{A^l}_{s\star}, \Sigmatilde^{A^l}_{s\star\star} \right) \,.
\end{equation}
A graphical model corresponding to this variational posterior is shown in \cref{fig:gpn_vi_distrs}b.
Note that this differs from the variational approximation employed for deep \Gps{}, which uses a \emph{mean-field} approach~\cite{damianou2013deep} that \emph{factorizes} over the variables $X^l$, $l \in \{1,2, \dots, L\}$, and is shown in \cref{fig:gpn_vi_distrs}c.

The variational parameters $\muhat^{U^l}$ and $\Sigmahat^{U^l}$ are estimated by maximizing the \ELBO{} given by
\begin{align}\label{eq:gpn_vi_sampling_l}
\L = -\idotsint & \Q(\{U\}_1^L, \{X\}_1^{L-1}, \{A\}_1^L, \{F\}_1^L)  \, \log \frac{\Q(\{U\}_1^L, \{X\}_1^{L-1}, \{A\}_1^L, \{F\}_1^L) }{\P(\{U\}_1^L, \{X\}_1^L, \{A\}_1^L, \{F\}_1^L \| X^0)} \, \cdot \nonumber \\
& \d\{U\}_1^L \, \d \{X\}_1^{L-1} \, \d \{A\}_1^L \, \d \{F\}_1^L \,.
\end{align}
Substituting the distributions into this equation results in $\L = -\L_{\mathrm{reg}} + \L_{\mathrm{pred}}$ with the following terms,
\begin{equation}\label{eq:gpn_vi_l}
\L_{\mathrm{reg}} = \sum_{l=1}^{L} \int \Q(U^l) \, \log \frac{\Q(U^l)}{\P(U^l)} \, \d U^l  \,,  
\quad\quad
\L_{\mathrm{pred}} = \int \Q(F^L) \, \log \P(X^L \| F^L) \, \d F^L   \,. 
\end{equation}
The term $\L_{\mathrm{reg}}$ can be identified as the sum of the KL-divergences of the virtual observation targets between their prior and variational posterior.
Its purpose is to keep the approximative posterior close to the prior and thus it can be understood as a regularization term.
Disregarding an additive constant, its value is given by
\begin{equation}\label{eq:gpn_vi_sampling_la}
\L_{\mathrm{reg}} \propto \frac{1}{2} \sum_{l=1}^{L} \sum_{n=1}^{N_l} \left( \tr\!\left( K^l_n(V^l_{\star n}, V^l_{\star n})^{-1} \, \Sigmahat^{U^l}_{\star\star n} \right) + (\muhat^{U^l}_{\star n})^T K^l_n(V^l_{\star n}, V^l_{\star n})^{-1} \muhat^{U^l}_{\star n} + \log \frac{\abs{K^l_n(V^l_{\star n}, V^l_{\star n})}}{\abs{\Sigmahat^{U^l}_{\star\star n}}} \right) \,. \
\end{equation}
For the term $\L_{\mathrm{pred}}$ we expand $\log \P(X^L \| F^L)$ over the samples and obtain, 
\begin{equation}\label{eq:gpn_lpred_exp}
\L_{\mathrm{pred}} = \sum_{s=1}^S \E_{\Q(F^L_{s\star} \| X^0_{s\star})}\!\left[ \log \P(X^L_{s\star} \| F^L_{s\star}) \right] \,, 
\end{equation}
and thus we can evaluate it like~\eqref{eq:gpn_parametric_regression_log_loss} by iterative propagation of moments from layer to layer as we will show now.
For $l \geq 1$ we have
\begin{equation}\label{eq:gpn_q_alp1}
\Q(A^{l+1}) = \iiint \Q(A^l) \, \Q(F^l \| A^l) \, \P(X^l \| F^l) \, \P(A^{l+1} \| X^l) \, \d A^l \, \d F^l \, \d X^l 
\end{equation}
with
\begin{equation}
\Q(F^l \| A^l) = \prod_{n=1}^{N_l} \int \Q(U^l_{\star n}) \, \P(F^l_{\star n} \| A^l_{\star n}, U^l_{\star n}) \, \d U^l_{\star n}  \,. \,. \label{eq:gpn_vi_sampling_fl_al}
\end{equation}
Since $\Q(F^l \| A^l)$ is the conditional of a \Gp{} with normally distributed observations, the joint distribution $\Q(F^l_{\star n}, U^l_{\star n} \| A^l_{\star n}) = \Q(U^l_{\star n}) \, \P(F^l_{\star n} \| A^l_{\star n}, U^l_{\star n})$ must itself be normal,
\begin{align}
\Q(F^l_{\star n}, U^l_{\star n} \| A^l_{\star n}) = 
\N\!\left( \begin{bmatrix} F^l_{\star n} \\ U^l_{\star n} \end{bmatrix} \,\Bigg|\,
\begin{bmatrix} \muhat^{F^l}_{\star n} \\ \muhat^{U^l}_{\star n} \end{bmatrix},
\begin{bmatrix} \Sigmahat^{F^l}_{\star\star n} & \Sigmatilde_{FU} \\
                \Sigmatilde_{UF} & \Sigmahat^{U^l}_{\star\star n} \end{bmatrix} \right) \,,
\end{align}
and we can find the values for the unknown parameters $\muhat^{F^l}_{\star n}$, $\Sigmahat^{F^l}_{\star\star n}$ and $\Sigmatilde_{FU} = \Sigmatilde_{UF}^T$ by equating the moments of its conditional distribution $\Q(F^l_{\star n} \| U^l_{\star n}, A^l_{\star n})$ with $\P(F^l_{\star n} \|  U^l_{\star n}, A^l_{\star n})$.
The conditional distribution is given by
\begin{equation}
\Q(F^l_{\star n} \| U^l_{\star n}, A^l_{\star n}) = \N(F^l_{\star n} \| \mutilde, \Sigmatilde) 
\end{equation}
and thus we obtain the following set of equations by comparing means and covariances,
\begin{align*}
\mutilde &\teq \muhat^{F^l}_{\star n} + \Sigmatilde_{FU} \, (\Sigmahat^{U^l}_{\star\star n})^{-1} \, (U^l_{\star n} - \muhat^{U^l}_{\star n}) 
= K^l_n(A^l_{\star n}, V^l_{\star n}) \, K^l_n(V^l_{\star n}, V^l_{\star n})^{-1} \, U^l_{\star n} \,,  \\
\Sigmatilde &\teq \Sigmahat^{F^l}_{\star\star n} - \Sigmatilde_{FU} \, (\Sigmahat^{U^l}_{\star\star n})^{-1} \, \Sigmatilde_{UF}
= K^l_n(A^l_{\star n}, A^l_{\star n}) - K^l_n(A^l_{\star n}, V^l_{\star n}) \, K^l_n(V^l_{\star n}, V^l_{\star n})^{-1} \, K^l_n(V^l_{\star n}, A^l_{\star n}) \,, \nonumber
\end{align*}
where the right sides are obtained from $\mu^{F^l}_{\star n}$, $\Sigma^{F^l}_{\star\star n}$ given by~\eqref{eq:gpn_parametrized_mu_sigma_y_sn} with $S^l_{\star n} = \vec{0}$ and without the additional variance $(\sigma^l_n)^2$.
Solving for the three unknowns gives
\begin{subequations}\label{eq:gpn_vi_sampling_muhat_sigmahat_fl}
\begin{align}
\muhat^{F^l}_{\star n} &= K(A^l_{\star n}, V^l_{\star n}) \, K(V^l_{\star n}, V^l_{\star n})^{-1} \, \muhat^{U^l}_{\star n} \label{eq:gpn_vi_sampling_muhat_fl} \\
\Sigmahat^{F^l}_{\star\star n} &= K(A^l_{\star n}, A^l_{\star n}) - K(A^l_{\star n}, V^l_{\star n}) \, \Khat^{U^l}_{\star\star n} \, K(V^l_{\star n}, A^l_{\star n}) \label{eq:gpn_vi_sampling_sigmahat_fl} \\
\Sigmatilde_{FU} &= K(A^l_{\star n}, V^l_{\star n}) \, K(V^l_{\star n}, V^l_{\star n})^{-1} \, \Sigmahat^{U^l}_{\star\star n} = (\Sigmatilde_{UF})^T 
\end{align}
\end{subequations}
with
\begin{equation}
\Khat^{U^l}_{\star\star n} \teq K(V^l_{\star n}, V^l_{\star n})^{-1} - K(V^l_{\star n}, V^l_{\star n})^{-1} \, \Sigmahat^{U^l}_{\star\star n} \, K(V^l_{\star n}, V^l_{\star n})^{-1} \,.
\end{equation}
Thus we obtain 
\begin{equation}\label{eq:gpn_vi_sampling_qflal}
\Q(F^l \| A^l) = \prod_{n=1}^{N^l} \N(F^l_{\star n} \| \muhat^{F^l}_{\star n}, \Sigmahat^{F^l}_{\star\star n}) \,.
\end{equation}
At this point it is instructive to verify that the obtained mean and covariance are consistent with the deterministic case and with the \Gp{} prior.
For deterministic observations, that is $\Sigmahat^{U^l}_{\star\star n}=0$, we obtain $\Khat^{U^l}_{\star\star n} = K(V^l_{\star n}, V^l_{\star n})^{-1}$ and thus recover standard \Gp{} regression.
If $U^l$ follows its prior, that is $\muhat^{U^l}_{\star n}=\vec{0}$ and $\Sigmahat^{U^l}_{\star\star n} = K(V^l_{\star n}, V^l_{\star n})$, we obtain $\muhat^{F^l}_{\star n} = \vec{0}$, $\Khat^{U^l}_{\star\star n} = 0$ and thus recover the \Gp{} prior on $F^l$.
In that case the virtual observations behave as if they were not present.

To calculate the mean $\mutilde^{A^{l+1}}_{s\star}$ and covariance $\Sigmatilde^{A^{l+1}}_{s\star\star}$ of the marginal $\Q(A^{l+1}_{s\star\star})$ from~\eqref{eq:gpn_q_alp1} we apply the same steps as in \cref{sec:gpn_propagation} to distribution~\eqref{eq:gpn_vi_sampling_qflal}, resulting in
\begin{equation}\label{eq:gpn_vi_a} 
\mutilde^{A^{l+1}}_{s\star} = \mutilde^{X^{l}}_{s\star} \,  W^{l+1} \,, \quad\quad\quad
\Sigmatilde^{A^{l+1}}_{s\star\star} = (W^{l+1})^T \, \Sigmatilde^{X^{l}}_{s\star\star} \, W^{l+1} \,, 
\end{equation}
The mean of $X^l$ is given by
\begin{equation}\label{eq:gpn_vi_mu}
\mutilde^{X^{l}}_{sn} = \psi^l_{s \star n} \, \kappa^l_{\star\star n} \,  \muhat^{U^l}_{\star n}
\end{equation}
with $\kappa$ and $\psi$ given by \cref{eq:gpn_normal_alpha_kappa,eq:gpn_normal_E_alpha}.
For the diagonal of the covariance matrix of $X^l$ we obtain
\begin{align}
\Sigmatilde^{X^{l}}_{snn} = 1 - \tr\!\left[ \left( \Khat^{U^l}_{\star \star n} - \beta^l_{\star n} \, (\beta^l_{\star n})^T \right)  \Omega^l_{\star \star n} \right] - \tr\!\left( \psi^l_{\star n} (\psi^l_{\star n})^T \, \beta^l_{\star n} (\beta^l_{\star n})^T \right) + (\sigma^l_n)^2  \label{eq:gpn_vi_marginal_sigma_fl}
\end{align}
with $\beta^l_{\star n} \teq K^l_n(V^l_{\star n}, V^l_{\star n})^{-1} \, \muhat^{U^l}_{\star n}$ and $\Omega^l_{rtn}$ from~\eqref{eq:gpn_normal_omega}.
The off-diagonal elements of the covariance matrix evaluate to
\begin{align}
\Sigmatilde^{X^{l}}_{snm} = \sum_r \sum_t \beta^l_{rn} \beta^l_{tm} \, \Lambda^l_{srtnm} - \mutilde^{F^l}_{sn} \, \mutilde^{F^l}_{sm} \label{eq:gpn_vi_marginal_sigma2_fl}
\end{align}
with $\Lambda^l_{srtnm}$ given by~\eqref{eq:gpn_normal_lambda}.

We can now propagate the moments of the occurring distributions from layer to layer by iterative application of \cref{eq:gpn_vi_mu,eq:gpn_vi_marginal_sigma_fl,eq:gpn_vi_marginal_sigma2_fl,eq:gpn_vi_a} and by doing so we obtain $\mutilde^{X^L}$ and $\Sigmatilde^{X^L}$.
With these quantities evaluating the expectation in~\eqref{eq:gpn_lpred_exp} gives
\begin{equation}\label{eq:gpn_lpred_marginal_final}
\L_{\mathrm{pred}} = - S \sum_{n=1}^{N_L} \log \sigma_n^L - \frac{1}{2} \sum_{s=1}^S \sum_{n=1}^{N_L} \frac{(X^L_{sn})^2 - 2\,X^L_{sn}\,\E\!\left[ F^L_{sn} \right] + \E\!\left[ (F^L_{sn})^2 \right]}{(\sigma^L_n)^2} \,. 
\end{equation}
with 
\begin{equation}
\E_{\Q(F^L_{s\star} \| X^0_{s\star})}\!\left[ F^L_{sn} \right] = \mutilde^{X^L}_{sn} \,,
\quad\quad
\E_{\Q(F^L_{s\star} \| X^0_{s\star})}\!\left[ (F^L_{sn})^2 \right] = 1 - \tr\!\left[ \left( \Khat^{U^L}_{\star \star n} - \beta^L_{\star n} \, (\beta^L_{\star n})^T \right)  \Omega^L_{s \star \star n} \right] \,,
\end{equation}
where $\beta^L_{\star n} \teq K^L_n(V^L_{\star n}, V^L_{\star n})^{-1} \, \muhat^{U^L}_{\star n}$ and $\Omega$ from~\eqref{eq:gpn_normal_omega}.

This concludes the calculation of all terms of the variational lower bound~\eqref{eq:gpn_vi_sampling_l}.
The resulting variational objective is a fully deterministic function of the parameters.
Training of the model is performed by maximizing $\L=-\L_{\mathrm{reg}} + \L_{\mathrm{pred}}$, with $\L_{\mathrm{reg}}$ given by~\eqref{eq:gpn_vi_sampling_la} and $\L_{\mathrm{pred}}$ given by~\eqref{eq:gpn_lpred_marginal_final}, \wrt to the variational parameters $\muhat^{U^l}$, $\Sigmahat^{U^l}$ and the model parameters $\vec{\sigma}^l$, $\vec{\lambda}^l$, $W^l$.
This can be performed using any gradient-descent based algorithm in a mini-batch training routine.
Here we have assumed a regression problem, but for a classification task we can apply the unscented transform as described in \cref{sec:eval_loss} to evaluate the resulting $\L_{\mathrm{pred}}$ term.

\clearpage
\section{The Relation to Deep Gaussian Processes}\label{sec:gpn_vs_deep_gp}

Deep Gaussian processes~\cite{damianou2013deep} is a framework for hierarchical composition of \Gp{} functions.
Similar to the \GPN{} model, outputs of one \Gp{} are used as the input for another one; thus graphical models resembling the structure of a feed-forward neural network can be formed.
Deep \Gps{} also employ the variational sparse \Gp{} method using inducing points developed by \textcite{titsias2009variational} to make inference tractable.
However, as we will show now, \GPNs{} have a number of advantages over deep \Gps{} both in model complexity and efficiency of inference.

A \Gp{} within the deep \Gp{} framework takes multidimensional input, \ie each input connection adds an input dimension to the \Gp{} it connects to.
The connections in a deep \Gp{} do not use weights to compute a weighted sum as it is done in the \GPN{} model.
Instead, each \Gp{} uses the \Ard{} covariance function that has an individual lengthscale parameter per input dimension.
By interpreting the lengthscale of the \ac{ARD} covariance function as the inverse of a weight, we can write for the covariance function of a deep \Gp{},
\[ k_{\mathrm{ARD}}(\vec{y}, \vec{y'}) = \exp\!\left[- \sum_{d} w_d^2 (y_d - y'_d)^2 \right]  \,. \]
Compared to that, the effective covariance function of a \GPN{} is
\[ k_{\mathrm{GPN}}(\vec{y}, \vec{y'}) =  \exp\!\left[- \left( \sum_{d} w_d (y_d - y'_d) \right)^2 \right] \,. \]
Thus taking the square \emph{before or after} summation is what distinguishes \GPNs{} from deep \Gps{} in their essence.
Although at first glance this seems to be a rather small difference, it leads to a series of consequences that clearly distinguishes both models.

\begin{figure}[tb]
\centering
\begin{subfigure}[b]{0.32\columnwidth}\centering
\caption{\Gp{} \Ard{}}
\includegraphics[width=0.99\columnwidth]{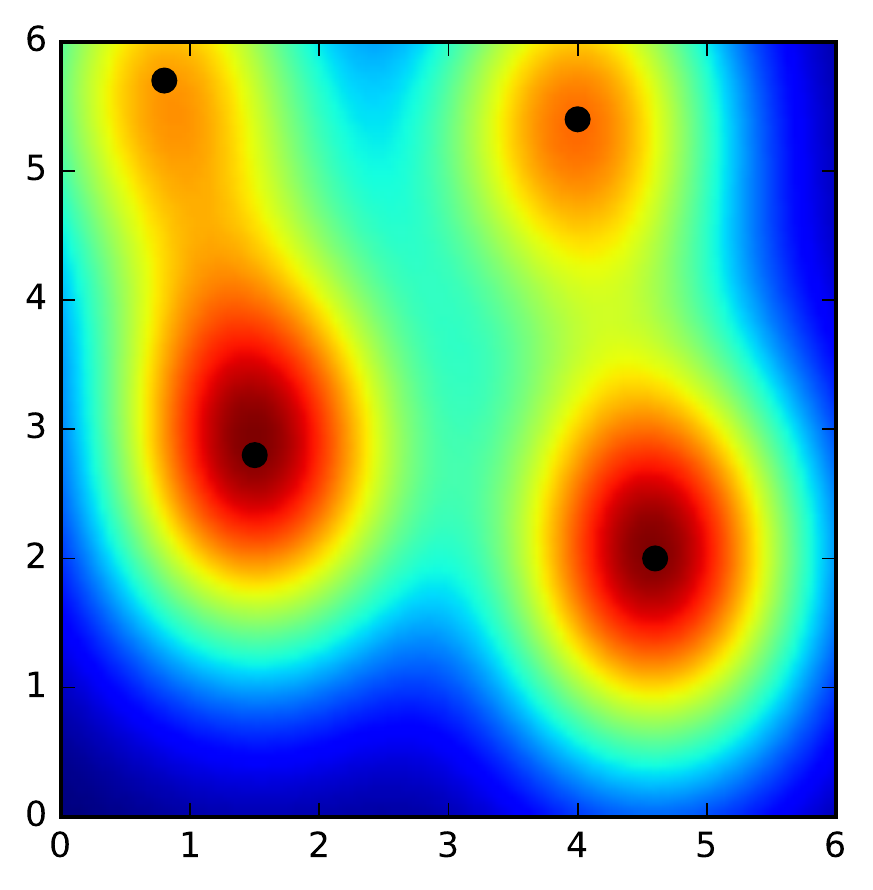}
\end{subfigure}\\  
\begin{subfigure}[b]{0.32\columnwidth}\centering
\includegraphics[width=0.99\columnwidth]{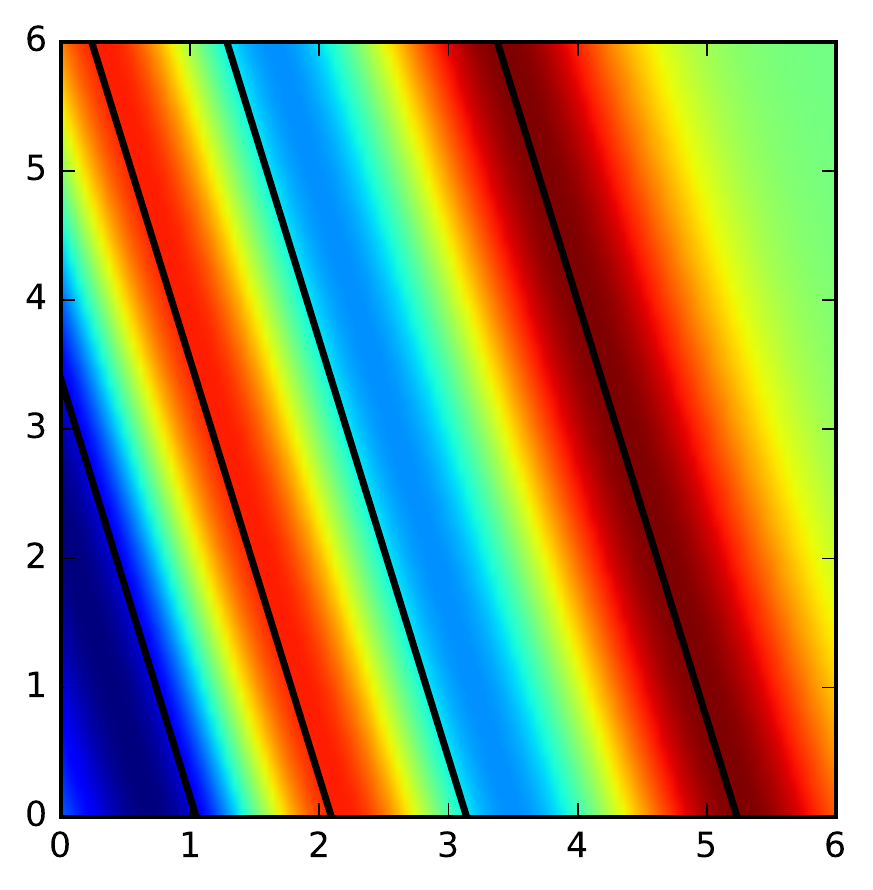}
\caption{\GPN{} 1}
\end{subfigure}
\begin{subfigure}[b]{0.32\columnwidth}\centering
\includegraphics[width=0.99\columnwidth]{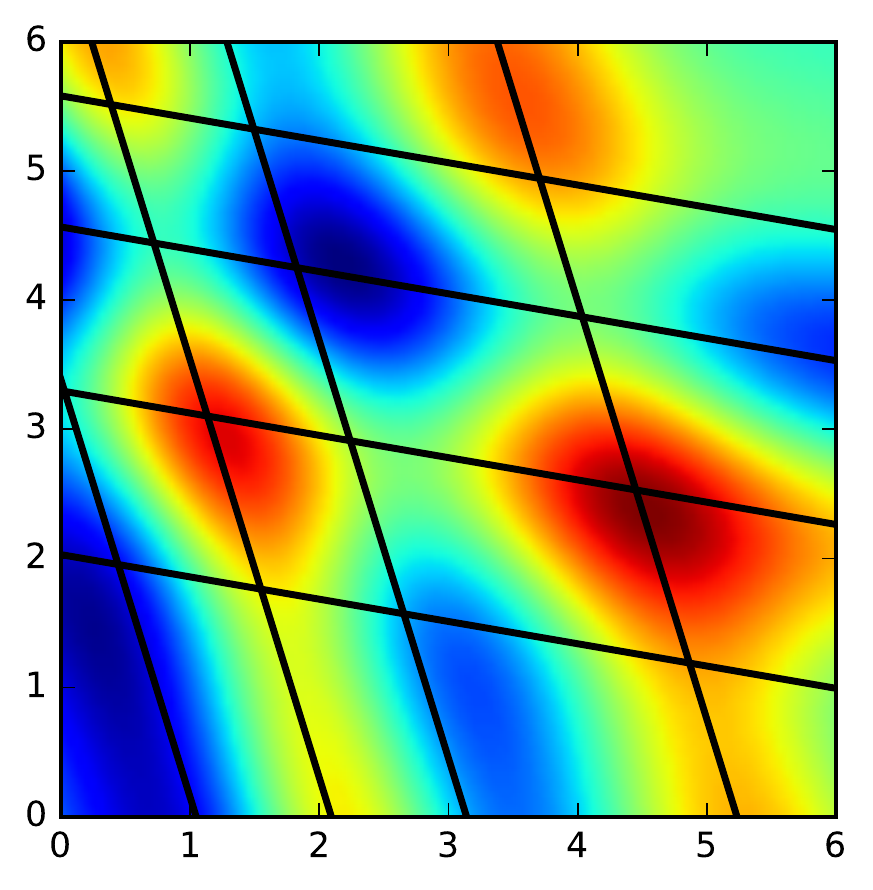}  
\caption{\GPN{} 1 + \GPN{} 2}
\end{subfigure}  
\begin{subfigure}[b]{0.32\columnwidth}\centering
\includegraphics[width=0.99\columnwidth]{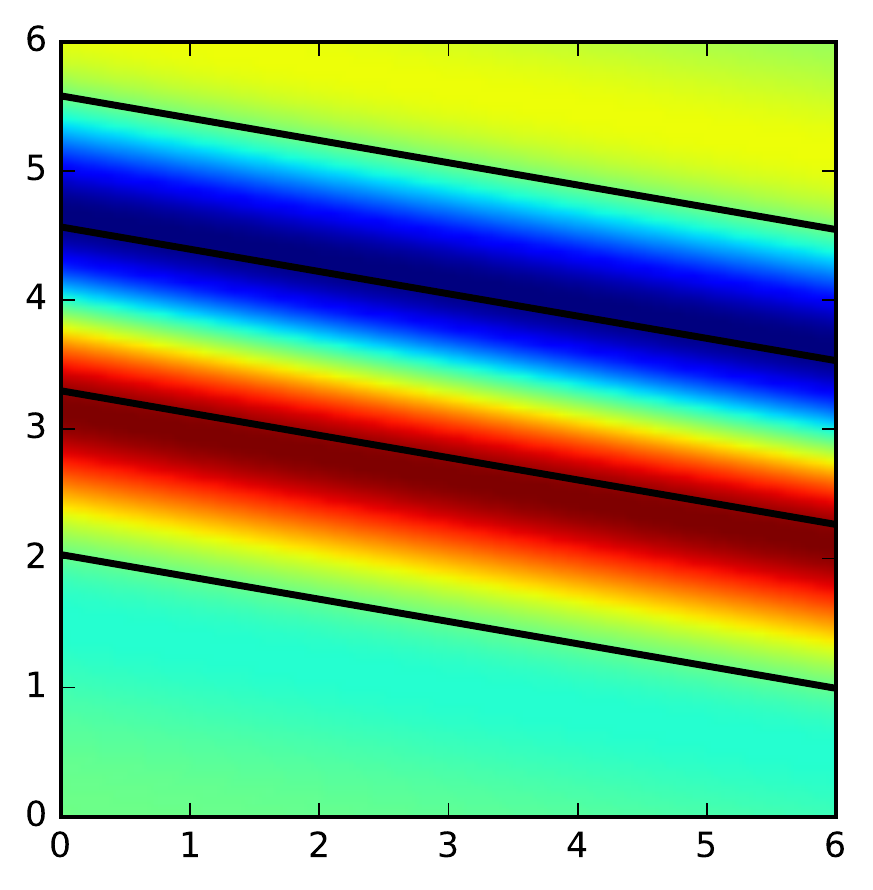}  
\caption{\GPN{} 2}
\end{subfigure}  
\caption{\Gp{} with \Ard{} covariance function as it occurs in a deep \Gp{} (top) versus \GPNs{} using projections (bottom) in two-dimensional space.
(a) The predictive mean of a \Gp{} in two-dimensional space using the \Ard{} covariance function and four observations placed in two-dimensional space.
(b) The predictive mean of a \Gp{} using a projection of two-dimensional inputs and four one-dimensional observations that are represented as lines in input space.
(d) Shows the same as (b) but using a different projection.
(c) A linear combination of the predictive means of (b) and (d) as it occurs in the activation of a \GPN{} receiving inputs from the \GPNs{} shown in (b) and (d).
Taken together their observations form a grid in two-dimensional space.
}
\label{fig:gpn_vs_ard}
\end{figure}

The first consequence is that each \Gp{} in the deep \Gp{} framework works in a \emph{multidimensional} function space.
The dimensionality is determined by the number of input connections and thus in a feed-forward model it equals the number of units in the previous layer.
Hence the inducing points of the virtual observations used for efficient inference are also multidimensional.
This implies that the number of virtual observations required to evenly cover the input space scales \emph{exponentially} with the number of input dimensions and thus incoming connections.
\Cref{fig:gpn_vs_ard}a shows the predictive mean of a two-dimensional \Gp{} with the \Ard{} covariance function with four observations.
As one moves further away from these observations the predictive mean returns to zero.

On the other hand a \GPN{}, like every artificial neuron, computes the projection of its inputs onto its weight vector resulting in a \emph{scalar} value.
Thus no matter how many input connections are present, a \GPN{} always works in a \emph{one-dimensional} function space.
Hence the inducing points of the virtual observations are also one-dimensional and the number of virtual observations per \GPN{} is unaffected by the number of incoming connections.
\Cref{fig:gpn_vs_ard}b shows the predictive mean of a \Gp{} using a projection of a two-dimensional input space and four observations.
Thus inducing points become inducing lines or inducing hyperplanes when more than two dimensions are concerned.
It might be argued that the expressive power is vastly reduced by using a projection, however this is not the case in a \GPN{} feed-forward network as the following argument demonstrates.
\Cref{fig:gpn_vs_ard}d also shows the predictive mean of a \Gp{} using a projection of a two-dimensional input space but with different weights.
Assume that \cref{fig:gpn_vs_ard}b and \cref{fig:gpn_vs_ard}d are the outputs of two \GPNs{} located in the same layer.
For the sake of argument further assume that this particular layer consists only of these two \GPNs{}.
Then the activations of a \GPN{} in the subsequent layer is formed by a linear combination of the output of these two \GPNs{}.
The resulting activation (using equal weights) is shown in \cref{fig:gpn_vs_ard}c and, as it can be seen, it produces functions varying in \emph{both} input dimensions.
Here, the number of virtual observations required to evenly cover the input space scales \emph{linearly} with the number of input dimensions.
Furthermore, the virtual observations can now be interpreted as a grid in input space, thus making it unlikely that an input point is located far away from all inducing hyperplanes.

The second consequence is that the inputs to a \Gp{} in a deep \Gp{} model cannot converge to a normal distribution because no linear combination (as in a neural network) is performed.
This leaves two methods for training and inference of a deep \Gp{}: stochastic variational inference, which is computationally expensive due to sampling, or using a \emph{mean-field} variational posterior (\cref{fig:gpn_vi_distrs}c) which does not work well in practice.
Furthermore, the mean-field outputs of each \Gp{} within a deep \Gp{} must be inferred alongside the model parameters during minimization of the variational objective function, leading to many more parameters to optimize.
For instance \textcite{salimbeni2017doubly} observed that deep \Gps{} are difficult to train for these reasons and reverted to a stochastic inference algorithm to avoid this problem, albeit at significantly higher computational costs.

On the other hand, the central limit theorem is applicable to the activation of each \GPN{}, guaranteeing that the activations will converge to a normal distribution if a \GPN{} has a sufficient number of input connection.
This leads to the \emph{marginally} normal variational posterior that we derived in \cref{sec:gpn_bayesian}.
The variational objective function resembles the structure of a neural network, which makes \GPNs{} directly usable in other network architectures such as \RNNs{} and \CNNs{} simply by adjusting \cref{eq:gpn_vi_a} accordingly.

Taken together both consequences show that the design of a \GPN{} leads to a more sound and efficient training procedure with fewer parameters to optimize compared to a deep \Gp{}.
These advances were possible by using a weight projection inside a standard \Se{} covariance function instead of an \Ard{} covariance function.
This crucial difference is what enabled us to derive the efficient variational objective.

\clearpage
\section{Benchmarks and Experiments}

This section describes the experiments performed to estimate the performance of \GPN{} feed-forward networks.
The datasets on which the models are evaluated are introduced and the performance is compared to conventional feed-forward neural networks.
Experiments are performed using maximum likelihood inference using the parametric \GPN{} model.
Later we will add experimental results for variational Bayesian training of the \GPN{} model.
Furthermore, this section discusses the empirical computational and memory requirements.

\subsection{Datasets}\label{sec:gpn_experiments_datasets}
To evaluate how well a parametric \GPN{} model performs on real-world classification problems, we test it on three datasets from the UCI Machine Learning Repository~\cite{UCI} as well as on the MNIST database of handwritten digits~\cite{MNIST}.
Hereby our aim is not yet to beat the current state of the art performance on these datasets, since the literature~\cite{agostinelli2014learning} shows that trainable activation function are mostly beneficial to large convolutional models for image classification.
Instead, we focus on verifying the implementation, efficiency and trade-offs of parametric \GPN{} feed-forward models on different kinds of classification datasets compared to conventional neural networks with a fixed sigmoidal or hyperbolic tangent activation function regularized by the Dropout technique~\cite{srivastava2014dropout}.
The datasets were primarily chosen so that they do not only differ in size but also cover different kinds of features and targets.
Consequently, successful training on this selection of datasets shows that \GPNs{} are applicable to a variety of tasks and it is worthwhile to implement convolutional architectures based on \GPNs{} to tackle current image classification problems on large datasets such as CIFAR-100~\cite{krizhevsky2009learning} and ImageNet~\cite{deng2009imagenet}.
We shortly describe the properties of the dataset before continuing with the training procedures.

\subsubsection*{UCI Letter Recognition Dataset}\label{sec:gpn_experiments_letterrec}
The UCI Letter Recognition dataset, first used in \cite{frey1991letter}, consists of $20\,000$~samples with 16~continuous input features per sample.
All features are calculated from pixel images of the 26~capital characters of the English alphabet, with each image showing a single letter in one of 20~different fonts.
Furthermore the character images are randomly distorted to increase the variation of the dataset.
The 16 precomputed features, consisting of statistical moments and edge counts, are used as input to the classifier.
The objective is to identify the character.

\subsubsection*{UCI Adult Dataset}\label{sec:gpn_experiments_adult}
The UCI Adult dataset, introduced by \cite{Kohavi:1996:SUA:3001460.3001502}, consists of 6~continuous and 8~categorical features containing census data taken from $48\,848$~U.S.\ citizens collected in the year~1994.
The continuous features consists amongst others of the age, weight, work hours per week and years of educations.
The categorical features include such information as highest obtained degree, martial status, race, sex and country of origin.
The binary target objective is to predict whether a person's income exceeded $50\,000$~USD or not.
This dataset contains missing features for some samples, which we replaced by an additional ``unknown'' category for categorical features and by zero for continuous features.

\subsubsection*{UCI Connect-4 Dataset}\label{sec:gpn_experiments_connect4}
In contrast to the previous datasets, the Connect-4 dataset (John Tromp, \textcite{UCI}) consists only of categorical features.
Each of its 42~features represents the state of a field on the board of a Connect-4 game with a board size of $6 \times 7$.
The categories encode whether the position is currently occupied by player 1, by player 2 or is free.
The dataset contains all legal positions in which neither player has won yet and in which the next move is not forced; in total the dataset contains $67\,557$~samples.
The data is used to classify the game result for player 1 if she plays optimally into one of the three classes: ``win'', ``loss'' or ``draw''.

\subsubsection*{MNIST Dataset}\label{sec:gpn_experiments_mnist}
The MNIST database of handwritten digits~\cite{MNIST} is one of the most commonly used machine learning datasets for image classification.
It consists of $60\,000$~training and $10\,000$~test examples.
We further split the training examples into a training and validation set.
Each input consists a $28 \times 28$ pixel image of a handwritten digit that has to be classified.
This task is similar to classification on the letter-recognition dataset, with the main difference being, that instead of using precomputed image features the model receives the raw, grayscale image data as input.
The MNIST dataset has been widely used to evaluate the performance of neural network based classifiers~\cite{hinton2006reducing,hinton2012improving,salakhutdinov2009deep,hinton2007recognize} and is thus a natural choice for evaluating trainable activation functions in a neural architecture.

\subsubsection{Preprocessing}
The continuous input features of all datasets are rescaled and shifted to lie in the interval $[0,1]$; for categorical features the one-hot encoding scheme is used.
It encodes a categorical feature as a vector having as many entries as there are categories with the entry for the active category being set to one and all other entries being set to zero.
The split between training and test set is kept as provided in the datasets; furthermore the original training set is randomly split into a smaller training set and a validation set consisting of 10\% of the original training samples.
The test set is only used to report final classification accuracies and not used in any way during training.

\subsection{Maximum Likelihood Training of Parametric GPNs}
In this section we perform experiments using the parametric \GPN{} with central limit activations developed in \cref{sec:parametric_gpn} with the goal of evaluating the feasibility of using \GPNs{} as a drop-in replacement for conventional artificial neurons.

\subsubsection{Model Variants}
As stated in theoretical analysis of the computational complexity in \cref{tab:gpn_normal_complexity}, propagating mean, variance and the full covariance matrix from layer to layer comes with different computational and memory requirements. 
To analyze the trade-offs between runtime and prediction accuracy of these different approximations we train parametric \GPNs{} by only propagating the mean, by propagating the mean and only the diagonal of the covariance matrix and by propagating the mean and the full covariance matrix.
To only propagate the mean, all layer variances $\Sigmatilde^{X^l}_{s\star\star}$ are assumed to be zero and only \cref{eq:gpn_normal_a,eq:gpn_normal_mu} are evaluated for each layer; thereby the probabilistic nature of the model is eliminated and the \GPNs{} become conventional neurons with an activation function that is given by interpolating between their inducing points and targets.
Including the variance $\diag(\Sigmatilde^{X^l}_{s\star\star})$ in the computations is done by setting all off-diagonal elements of the layer covariance matrices to zero and using \cref{eq:gpn_normal_a,eq:gpn_normal_Sigmatilde}.
For the full model we propagate $\Sigmatilde^{X^l}_{s\star\star}$ from layer to layer by employing \cref{eq:gpn_normal_a,eq:gpn_normal_Sigmatilde_cov}.
The wall clock time and memory requirements of each approach are measured.

In preparatory experiments it became apparent that the virtual observation variances were driven to zero, leading to overfitting.
Such problems are common with maximum likelihood inference and thus a penalty of the form
\begin{equation}\label{eq:gpn_ls}
\mathscr{L}_{S}(\theta) = \sum_{l=1}^L \frac{1}{N_l} \sum_{n=1}^{N_l} \frac{1}{R} \sum_{r=1}^R \alpha_S \, \sigma\!\left( \frac{\beta_S}{\abs{S^l_{rn}}} \right) 
\end{equation}
where $\sigma$ is the logistic function, $\alpha_s = 0.1$ and $\beta_s = 10^{-3}$, was added to the loss function.
Note that this ad-hoc penalty is unnecessary when the variational Bayesian training objective derived in \cref{sec:gpn_bayesian} is used.

The virtual observations of the parametric \GPN{} model can be shared between different \GPNs{} resulting in \GPNs{} that use the same activation function.
Obviously this reduces the number of model parameters and furthermore the computational complexity.
To assess the impact of sharing on model accuracy, we train two variants of \GPN{} feed-forward networks: an independent, where each \GPN{} has its individual virtual observations, and a layer-shared, where all \GPNs{} within a layer share one activation function.

\subsubsection{Implementation}
Implementation was performed in a custom framework that generates CUDA kernels for complex expressions like \cref{eq:gpn_normal_E_alpha,eq:gpn_normal_omega,eq:gpn_normal_Sigmatilde_cov}.
Contrary to established frameworks like Theano \cite{theano} or TensorFlow~\cite{tensorflow}, such an expression and its derivatives are evaluated fully inside a \emph{single} CUDA kernel (including the occurring sums).
This proved essential to obtain acceptable performance, which is otherwise hampered by the occurrence of many small tensors corresponding to the virtual observations in the computational graph.
Derivatives of such expressions were calculated using the method described in~\cite{urban2017automatic}.

\subsubsection{Initialization}
Preparatory experiments showed that the inducing points $V^l_{rn}$ of each \GPN{} remained mostly unchanged during training; hence 14 inducing points are initialized using linear spacing in the interval $[-2, 2]$ and kept fixed during training.
The corresponding targets $U^l_{rn}$ are either initialized from a standard normal distribution or set equal to $V^l_{rn}$, resulting in the identity function.
We also tried initializing the targets to values of a well-known activation function, such as the hyperbolic tangent or the rectifier, but found no significant benefit.
All virtual observation variances $S^l_{rn}$ are initialized to the constant value of $\sqrt{0.1}$ and optimized during training alongside with the targets.

The weights $W^l_{nm}$ of each layer $l$ are initialized using a uniform distribution with support $[-r, r]$ where $r = \sqrt{6} / \sqrt{N_{l-1} + N_{l+1}}$.
This initialization has been recommended by \cite{glorot2010understanding} for training of deep neural network using the hyperbolic tangent activation function.
The motivation behind choosing $r$ as described is to ensure that at the beginning of training the activations of most neurons start in the linear range of the hyperbolic tangent function.
Although we are not using this function, it is desirable for the activations of \GPNs{} to fall within the range of their inducing points; thus this weight initialization method is applicable here.

\subsubsection{Training}
Training is performed by minimizing the expected loss $\mathscr{L}(\theta)$ as calculated by the unscented transform~\eqref{eq:gpn_normal_unscented} of the softmax cross-entropy loss~\eqref{eq:gpn_parametrized_loss_classification} using the Adam optimizer \cite{kingma2014adam}.
The initial learning rate is $10^{-3}$ and is decreased by factor 10 each time the validation loss stops improving for a predefined number of training iterations.
When the learning rate reaches $10^{-6}$ and no improvement is seen on the validation set, training is terminated and the model parameters of the best iteration as seen on the validation set are used to calculated the reported classification accuracies.
Each experiment is repeated five times with different random seeds for the initialization.
For comparison we also train a conventional neural network of the same architecture with a fixed hyperbolic tangent activation function and regularized using the fast Dropout method~\cite{Wang:2013}.

\subsubsection{Results}
\begin{figure}
\centering
\includegraphics[width=0.7\linewidth]{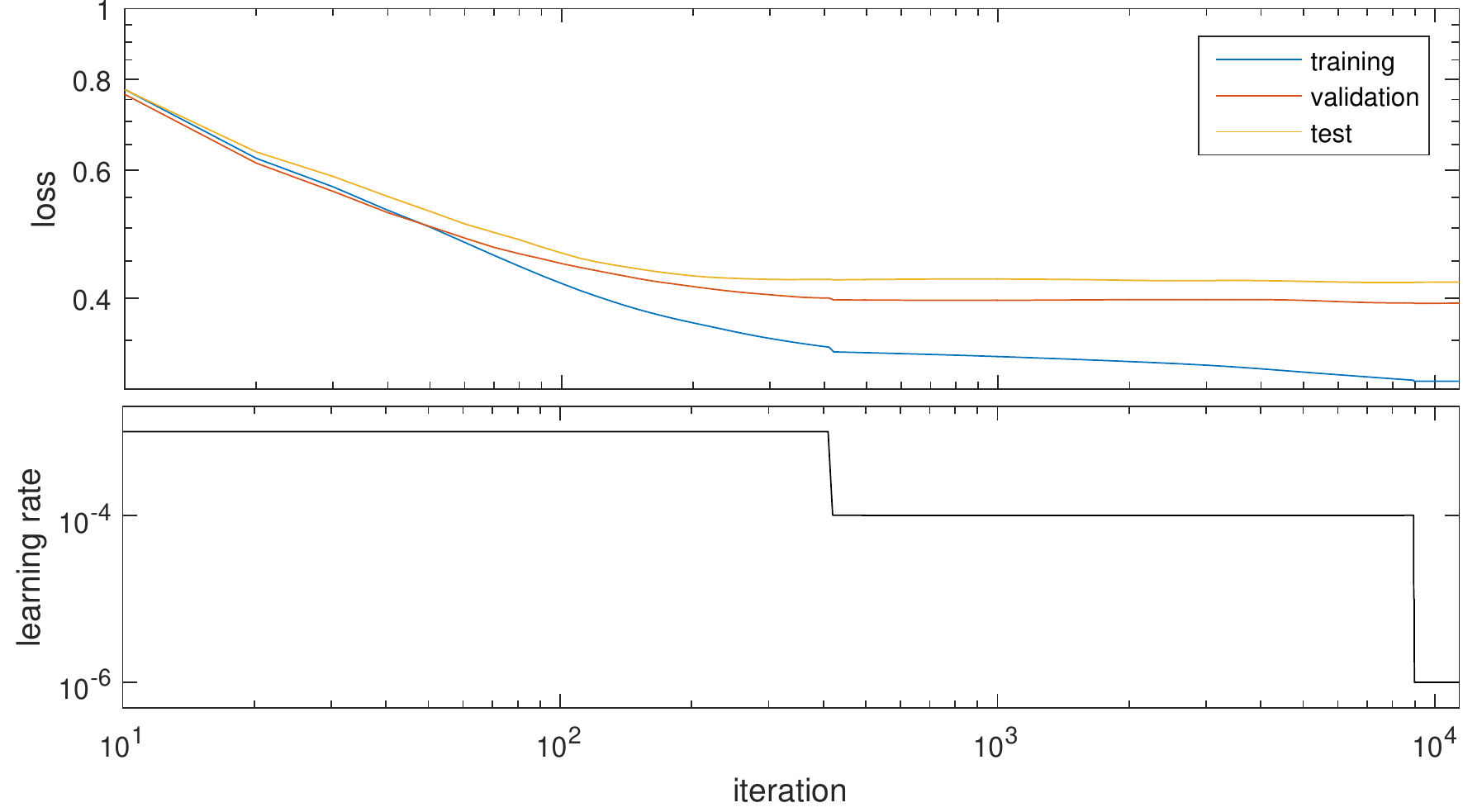}
\caption{
Training, validation and test losses during training of a parametric \GPN{} feed-forward network with mean and variance propagation on the UCL Connect-4 dataset.
The lower panel shows the scheduling of the learning rate.
The progression of the loss is smooth and stable due to the use of a fully deterministic objective.
}
\label{fig:gpn_loss_connect4}
\end{figure}

The below results are preliminary, as we did no hyperparameter tuning within the GPN architecture.
An exemplary loss curve and training rate schedule of a parametric \GPN{} feed-forward network is shown in \cref{fig:gpn_loss_connect4}.
In this example the initial learning rate of $10^{-3}$ is automatically decreased after 400 iterations to $10^{-4}$ and then again after $9\,000$ iterations to $10^{-6}$.
Training is terminated after $20\,000$ iterations.
Like a conventional feed-forward neural network, the losses decrease smoothly due to the use of a fully deterministic loss.

The accuracies of all experiments are reported in \cref{tab:gpn_exp_errors} and the resource usage of the different propagation approaches is shown in \cref{tab:gpn_exp_mem_time}.
As expected the conventional neural network with a fixed activation function is fastest; however relative to that using a \GPN{} is only four times slower when propagating the means and the variances.
This includes both the times for forward propagation and for calculating the gradient \wrt the weights and virtual observations using back propagation.
The propagation of the \GPN{} mean and variance leads to significantly better results than the propagation of the mean alone on all datasets.
However, the propagation of the full covariance matrix is about 12 times computationally more expensive than the propagation of the mean and variance and it did not show significant benefits to the accuracy of the model in preparatory experiments.

Sharing the \GPN{} virtual observations over all \GPNs{} within a layer does not provide any benefits on the test accuracy in our experiments, thus suggesting that the flexibility of having a separate activation function per \GPN{} is beneficial to the model and the increase in the number of parameters does not lead to overfitting.
On the UCL Adult dataset \GPNs{} profited from initializing their virtual observations so that the initial activation function is the identity function; however doing the same on the UCL Letter Recognition dataset did not yield any significant improvement.

\begin{sidewaystable}\centering\footnotesize
\begin{tabular}{llllllll@{}l} \toprule 
\emph{Dataset} 		& \emph{Layer sizes} & \emph{Units} & \emph{Sharing} & \emph{Act. init.} & \emph{Train error} & \emph{Val. error} & \multicolumn{2}{l}{\emph{Test error}} \\ \midrule
UCL Letter Rec.   
& 16x30x15x26 & fixed tanh 			 	& & 		    & $0.0463 \pm 0.0039$ & $0.0690 \pm 0.0068$ & $0.0765$ & $\pm 0.0035$ \\
& 16x30x15x26 & fixed tanh (dropout) 	& &    	        & $0.0354 \pm 0.0054$ & $0.0603 \pm 0.0046$ & $\mathbf{0.0625}$ & $ \pm 0.0042$ \\
& 16x30x15x26 & \GPN{} mean only    	& none & random & $0.0696 \pm 0.0163$ & $0.0690 \pm 0.0138$ & $0.0765$ & $ \pm 0.0104$ \\
& 16x30x15x26 & \GPN{} mean + variance 	& none & random & $0.0366 \pm 0.0052$ & $0.0668 \pm 0.0061$ & $0.0709$ & $ \pm 0.0043$ \\
& 16x30x15x26 & \GPN{} mean + variance 	& none & identity&$0.0348 \pm 0.0043$ & $0.0675 \pm 0.0034$ & $0.0715$ & $ \pm 0.0041$ \\
& 16x30x15x26 & \GPN{} mean + variance 	& layer& random & $0.0410 \pm 0.0060$ & $0.0704 \pm 0.0026$ & $0.0733$ & $ \pm 0.0033$ \\
\hline
UCL Adult
& 104x30x15x2 & fixed tanh				& & 		    & $0.1417 \pm 0.0070$ & $0.1586 \pm 0.0028$ & $0.1561$ & $ \pm 0.0032$ \\
& 104x30x15x2 & fixed tanh (dropout)	& & 		    & $0.1431 \pm 0.0006$ & $0.1420 \pm 0.0015$ & $0.1418$ & $ \pm 0.0014$ \\
& 104x30x15x2 & \GPN{} mean + variance 	& none & random & $0.1413 \pm 0.0015$ & $0.1490 \pm 0.0051$ & $0.1514$ & $ \pm 0.0059$ \\
& 104x30x15x2 & \GPN{} mean + variance 	& none & identity&$0.1469 \pm 0.0003$ & $0.1408 \pm 0.0018$ & $\mathbf{0.1390}$ & $ \pm 0.0008$ \\
\hline
UCL Connect-4
& 126x30x15x3 & fixed tanh				& & 		    & $0.0129 \pm 0.0044$ & $0.1535 \pm 0.0014$ & $0.1637$ & $ \pm 0.0021$ \\
& 126x30x15x3 & fixed tanh (dropout)	& & 		    & $0.1372 \pm 0.0037$ & $0.1454 \pm 0.0017$ & $\mathbf{0.1568}$ & $ \pm 0.0019$ \\
& 126x30x15x3 & \GPN{} mean + variance	& none & random & $0.1222 \pm 0.0063$ & $0.1494 \pm 0.0048$ & $0.1608$ & $ \pm 0.0017$ \\
& 126x30x15x3 & \GPN{} mean + variance	& layer& random & $0.1273 \pm 0.0059$ & $0.1501 \pm 0.0054$ & $0.1617$ & $ \pm 0.0062$ \\
\hline
MNIST Digits
& 784x30x15x10 & fixed tanh				& & 		    & $0.0162 \pm 0.0023$ & $0.0451 \pm 0.0016$ & $0.0546$ & $ \pm 0.0021$ \\
& 784x30x15x10 & \GPN{} mean + variance	& none & random & $0.0212 \pm 0.0017$ & $0.0426 \pm 0.0009$ & $\mathbf{0.0521}$ & $ \pm 0.0016$ \\
\bottomrule 
\end{tabular}
\caption{
Misclassification rates of different parametric \GPN{} model variants on the benchmark datasets.
The error is calculated as the number of misclassified samples divided by the number of total samples.
The error rates are given as the average of five experiments and with a confidence interval of $68\%$.
For evaluation the model parameters at the training iteration with the lowest validation loss were used.
}
\label{tab:gpn_exp_errors}
\end{sidewaystable}

\begin{table}\centering
\begin{tabular}{lrr} \toprule
\emph{Propagation using} 		& \emph{Memory usage}  	& \emph{Iteration time} \\ \midrule
fixed act. function			    & $30$  MB 				& $9$ ms \\
\hline
\GPN{} means only 				& $94$  MB 				& $29$  ms \\
\GPN{} means and variances 		& $113$ MB 				& $36$  ms \\
\GPN{} means and full covariances & $227$ MB 			& $118$ ms \\
\bottomrule 
\end{tabular}
\caption{
Memory usage and time for performing one iteration of forward- and back-propagation of a layer of 50 neurons or \GPNs{} using different propagation methods.
Values should only be used for relative comparisons within this table since some irrelevant operations, such as data reading, are included in the memory usage and iterations time.
Memory usage includes the memory used for storing intermediate results for calculation of the derivatives using backpropagation.
}
\label{tab:gpn_exp_mem_time}
\end{table}

\begin{figure}[!tb]
\centering
\includegraphics[width=1.0\linewidth,trim={0 0 10mm 0},clip]{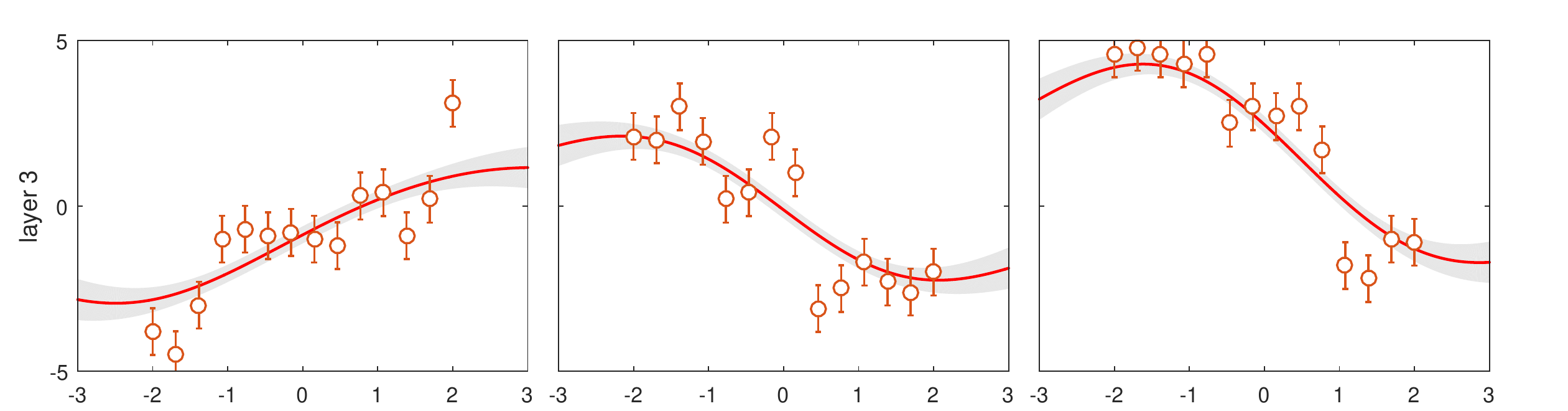} \\
\includegraphics[width=1.0\linewidth,trim={0 0 10mm 0},clip]{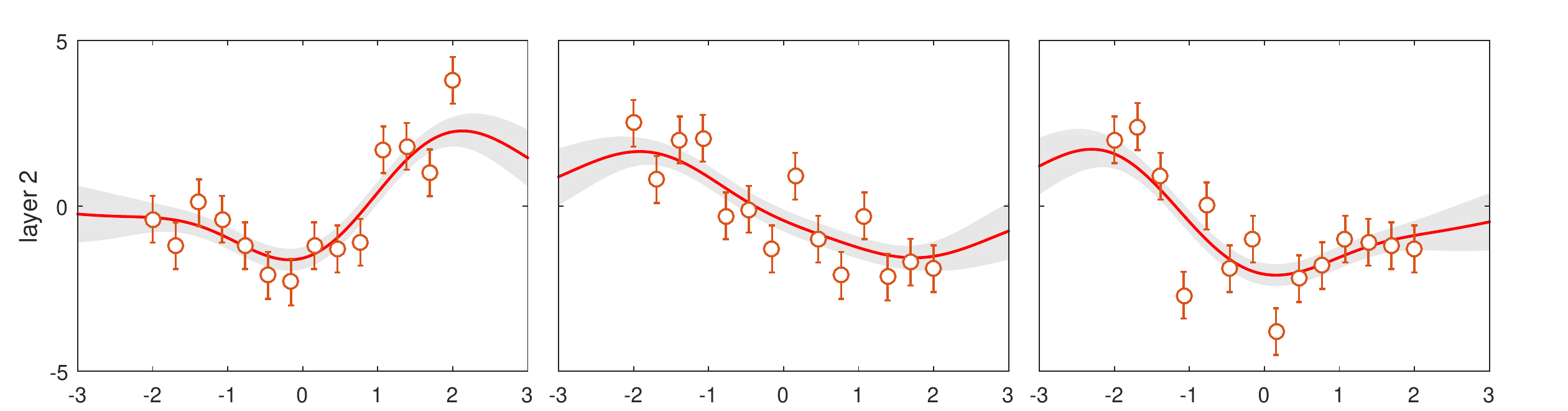} \\
\includegraphics[width=1.0\linewidth,trim={0 0 10mm 0},clip]{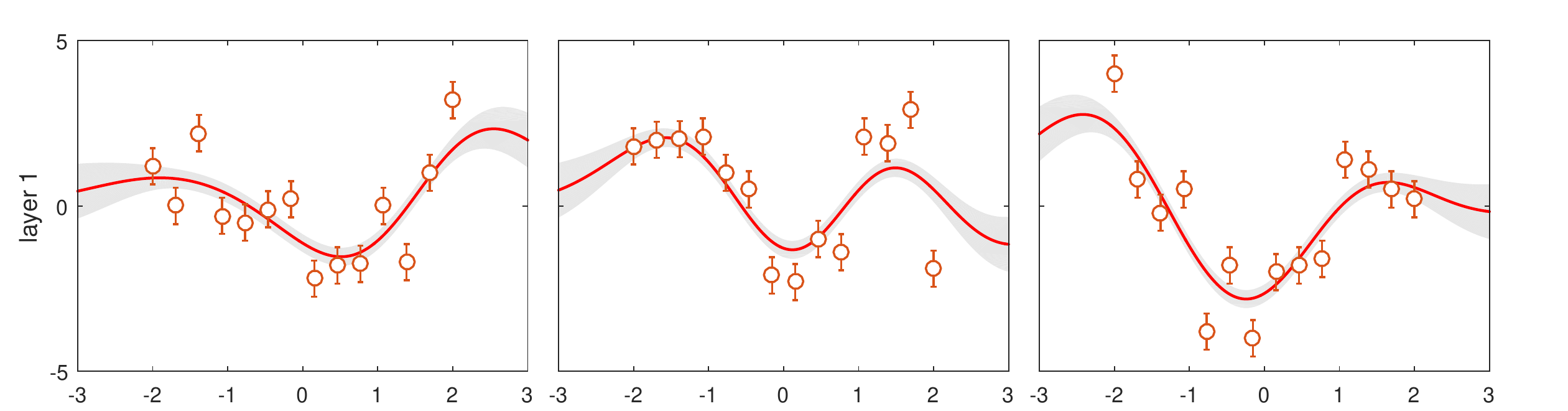} 
\caption{
Three activation function from each layer of a parametric \GPN{} feed-forward network that was trained on the UCL Connect-4 dataset.
The virtual observations are shown as red dots together with their standard deviation.
}
\label{fig:gpn_act_connect4}
\end{figure}

\Cref{fig:gpn_act_connect4} shows examples of activation functions that are commonly encountered in a \GPN{} feed-forward network  after training it on UCL Connect-4 dataset.
The activation functions in the first layer vary much stronger than those in the upper two layers.
Most commonly functions in the first layer resemble sine-like functions and are approximately axis-symmetric \wrt the y-axis.
As we move to the second and third layer, sigmoid-shaped and linear functions become more common.
This might indicate that the first layer exploits a periodicity in the input data while the top two layers act as feature-detectors by gating their inputs.

\subsubsection{Discussion}
The results presented here show that parametric \GPNs{} have consistently better performance than a conventional neural network using a fixed activation function on real-world datasets of small size.
The execution speed of a \GPN{} network is about 25\% of that of a conventional \ANN{} with a fixed activation function.
Propagating mean and variances through the \GPN{} network significantly improves the quality of its predictions and its generalization ability.
Propagating full covariance matrices yields only marginal improvements, which cannot be justified by the vastly increased usage of computational time and memory.
The benefit of how the activation functions are initialized is dataset-dependend and sharing virtual observations does not yield improvements.

When a conventional neural network is regularized using the fast Dropout method, \GPNs{} win on some datasets while conventional neural networks perform better on others.
We strongly suspect that the generalization capabilities of GPNs much outperform deep neural networks with, e.g., ReLU functions, due to stronger nonlinearities in their neurons.  Future experiments will be conducted to investigate this assumption, esp.\ related to CNNs.
Note that these experiments were performed on small datasets using maximum likelihood inference and \emph{not} the Bayesian inference described in \cref{sec:gpn_bayesian}.
Future experiments will concentrate on testing \GPNs{} in \CNNs{} using large datasets, because historically novel and learnable activation functions were shown to significantly improve~\cite{NIPS2012_4824,maas2013rectifier,agostinelli2014learning,he2015delving} these models.

\subsection{Bayesian Training of GPNs}
In this section we perform experiments using \GPN{} networks trained using approximative Bayesian inference~\cref{sec:gpn_bayesian}.

\textbf{Experiments are currently being conducted and results will be added when they become available.}

\clearpage
\section{Conclusion}

We proposed to place a \Gp{} prior over the activation function of each neuron.
This has three consequences.
First, the neuron using this activation function becomes a probabilistic unit, allowing it to handle uncertain inputs and estimate the confidence of its output.
Second, complexity of the activation function is penalized in a probabilistically sound Bayesian setting; this guards the model against overfitting.
Third, the squared exponential covariance function ensures that all activation functions are smooth and therefore continuous derivatives are available.
This resulted in the non-parametric \GPN{} model, which shows these theoretically attractive properties, but performing inference is expensive due to its non-parametric nature.

\begin{figure}
\centering
\scalebox{0.7}{
\definecolor{block_model_color}{rgb}{1.0,1.0,1.0}
\definecolor{block_concept_color}{rgb}{1.0,1.0,1.0}
\definecolor{block_method_color}{rgb}{1.0,1.0,1.0}
\definecolor{block_posterior_color}{rgb}{1.0,1.0,1.0}
\begin{tikzpicture}[auto,
  block_model/.style={rectangle,draw=black,thick,fill=block_model_color,text width=8em,text centered,minimum height=4em,text width=2cm,text=black,rounded corners=.15cm},
  block_concept/.style={rectangle,draw=black,thick,fill=block_concept_color,text width=16em,text centered,minimum height=3em,text width=2.2cm,text=black,rounded corners=.15cm},
  block_method/.style={rectangle,draw=black,thick,fill=block_method_color,text width=8em,text centered,minimum height=4em,text width=2cm,text=black,rounded corners=.15cm},
  block_posterior/.style={rectangle,draw=black,thick,fill=block_posterior_color,text width=8em,text centered,minimum height=3em,text width=2.9cm,text=black,rounded corners=.15cm},  
  line/.style ={draw, thick, -latex', shorten >=0pt}]
  
\tikzstyle{annot} = [text width=4cm, text centered]	  
  
\node[block_model] (GPN) at (0.0,0.0) {{\bf \large GPN}\\\small (sec.~\ref{sec:gpn})};
\node[block_model] (parametric GPN) at (4.0,-4.0) {{\bf parametric GPN}\\\small (sec.~\ref{sec:parametric_gpn})};
\node[block_model] (variational GPN) at (0.0,-8.0) {{\bf variational GPN}\\\small (sec.~\ref{sec:gpn_bayesian})};
\node[block_model] (normalized GPN) at (8.0,-8.0) {{\bf central limit GPN}\\\small (sec.~\ref{sec:gpn_propagation})};

\node[block_concept] (virtualobs) at (5.9,-2.0) {virtual\\observations};
\node[block_concept] (GP prior on virtual obs) at (2.0,-6.0) {GP prior on virtual obs.};
\node[block_concept] (CLT) at (4.0,-10.0) {central limit theorem};

\node[block_method] (SBT2) at (11.0,-4.0) {stochastic backprop training};
\node[block_method] (BT2) at (11.0,-8.0) {backprop training};

\node[block_posterior] (marginalized) at (-3.5,-8.0) {marginal posterior};

\node[block_method] (HMC) at (-7.0,0.0) {HMC training / inference};
\node[block_method] (BT) at  (-7.0,-8.0) {backprop training};

\draw[->,thick] (GPN.east) to [bend left=45] (parametric GPN.north);
\draw[thick] (virtualobs.west) to [bend right=25] (4.0009,-3.0287);
\draw[->,thick] (parametric GPN.south) to [bend left=45] (variational GPN.east);
\draw[->] (parametric GPN.east) to (SBT2.west);
\draw[thick] (GP prior on virtual obs.south) to [bend left=25] (1.5002,-7.9512);
\draw[->,thick] (parametric GPN.south) to [bend right=45] (normalized GPN.west);
\draw[->] (normalized GPN.east) to (BT2.west);
\draw[->,thick,to path={|- (\tikztotarget)}] (variational GPN.west) to (marginalized.east);

\draw[->,thick,to path={-| (\tikztotarget)}] (CLT.east) to (normalized GPN.south);
\draw[->,thick,to path={-| (\tikztotarget)}] (CLT.west) to (marginalized.south);

\draw[->] (GPN.west) to (HMC.east);
\draw[->] (marginalized.west) to (BT.east);
 
\node[annot] at (-7.0,-10.8) {\textit{Bayesian \\ inference}};
\node[annot] at (10.2,-10.8) {\textit{maximum-likelihood \\ inference}};

\end{tikzpicture}}
\caption{Overview of the family of \GPN{} models and their relationships.}
\label{fig:gpn_overview}
\end{figure}
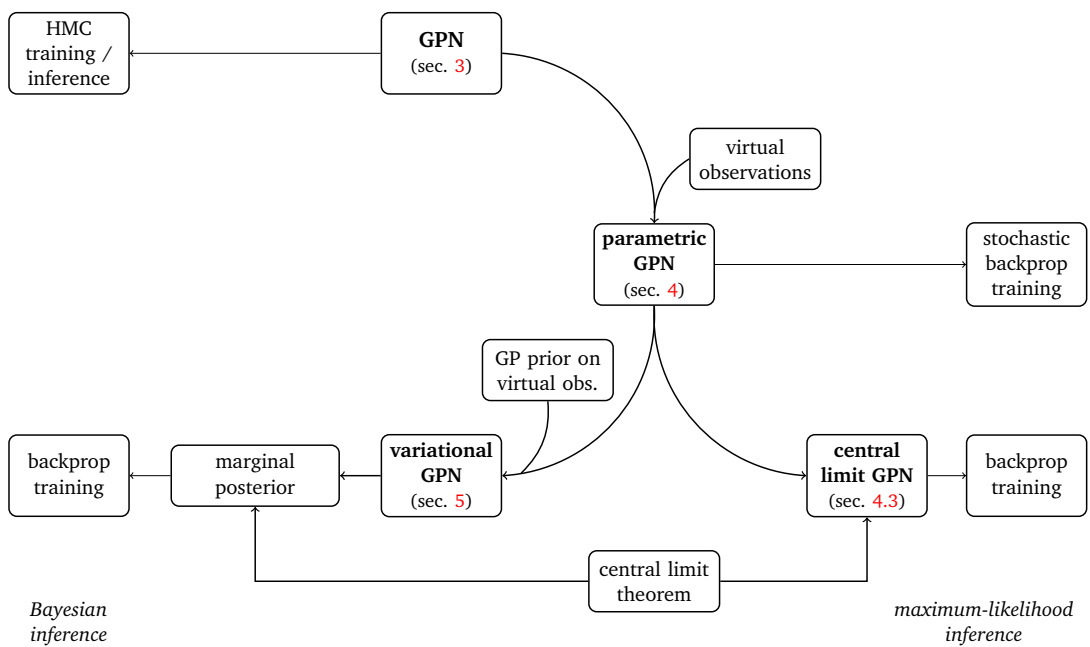

An overview of the course of action we took to make \GPNs{} tractable is shown in \cref{fig:gpn_overview}.
Starting from a non-parametric model we derived a variational approximation of the posterior.
Based on methods proposed for sparse \Gp{} regression, we introduced an auxiliary model, the parametric \GPN{}, that provides inexpensive inference but is also less attractive since inference is performed by maximizing the likelihood.
We then showed that it is possible to recover the non-parametric \GPN{} model by placing an appropriate prior over the parameters of a parametric \GPN{}.
Furthermore, we showed that the distribution of activations in both randomly initialized and fully trained neural networks closely resembles a normal distribution due to the central limit theorem.
Taken together, these two steps allowed us to derive a fully deterministic, variational objective to train a non-parametric \GPN{} by stochastic gradient descent.
This objective has the same functional structure as that of a conventional neural network and thus \GPNs{} can be directly included in \CNNs{} and \RNNs{} or any other architecture that uses neurons.

We have shown that, although \GPN{} networks are similar to deep \Gps{}, they need significantly fewer (variational) parameters and training is vastly more efficient.
This is the result of using projections inside the covariance function instead of dimension-dependent lengthscales.

Preliminary experimental results on small datasets using parametric \GPNs{} trained by maximizing the parameter likelihood show that the model performs consistently better than a conventional neural network with a fixed activation function.
Experiments on larger datasets and using approximate Bayesian inference are ongoing and results will be presented as they become available.

In summary, from a neural network viewpoint we have introduced a novel, stochastic, learnable, self-regularizing activation function that is integratable into existing neural models with modest effort.
From a \Gp{} viewpoint we introduced the idea of learnable projections into deep Gaussian processes, allowing us to derive a novel variational posterior that makes them as accessible and easy to train as neural networks.


\clearpage
\bibliographystyle{apalike}
\bibliography{gpn}

\end{document}